%% file: main_arxiv.tex
\documentclass[10pt,twocolumn,letterpaper]{article}

\usepackage{cvpr}              

\input{preamble}

\definecolor{cvprblue}{rgb}{0.21,0.49,0.74}
\usepackage[pagebackref,breaklinks,colorlinks,citecolor=cvprblue]{hyperref}
\usepackage{multirow}
\newcommand{\com}{\textit{complexity}\xspace}
\newcommand{\vul}{\textit{vulnerability}\xspace}
\newcommand{\Com}{\textit{Complexity}\xspace}
\newcommand{\Vul}{\textit{Vulnerability}\xspace}

\title{Anomaly Score: Evaluating Generative Models and Individual Generated Images based on Complexity and Vulnerability}

\author{Jaehui Hwang$^*$, Junghyuk Lee$^*$, Jong-Seok Lee
\\School of Integrated Technology, Yonsei University~~~\small{$^*$ Equal contribution}
\\{\tt\small \{jaehui.hwang, junghyuklee, jong-seok.lee\}@yonsei.ac.kr}
}

\begin{document}
\maketitle
\input{sec/0_abstract}    
\input{sec/1_intro}
\input{sec/2_relatedwork}
\input{sec/3_why}

\input{sec/4_metric}
\input{sec/5_individual_metric}
\input{sec/6_conclusion}

\clearpage

{
    \small
    \bibliographystyle{ieeenat_fullname}
    \bibliography{main}
}


\newpage

\onecolumn
\maketitlesupplementary
\appendix
\counterwithin{figure}{section}
\counterwithin{table}{section}

In this supplementary material, we include additional materials, which are not contained in the main paper because of the page limit, such as an explanation of the employed generative models and details of the linear regression on vulnerability. We also provide additional experimental results on various feature models and examples of generated images that are used for the subjective test.

\section{Employed generative models}
We utilize various generated datasets from \url{https://github.com/layer6ai-labs/dgm-eval} \cite{dgm-eval}, which are listed below with respect to the target image dataset.

\begin{itemize}
    \item CIFAR10 \cite{cifar}: ACGAN \cite{acgan}, BigGAN \cite{biggan}, IDDPM \cite{iddpm}, LoGAN \cite{logan}, LSGM \cite{lsgm}, MHGAN \cite{mhgan}, PFGM++ \cite{pfgm++}, ReacGAN \cite{reacgan}, ResFlow \cite{resflow}, StyleGAN-XL \cite{sgxl}, StyleGAN2-ada \cite{sg2}, WGAN \cite{wgan}
    \item ImageNet \cite{imagenet}: ADM \cite{adm}, BigGAN \cite{biggan}, DiT-XL-2 \cite{ditxl2}, GigaGAN \cite{gigagan}, LDM \cite{ldm}, Mask-GIT \cite{maskgit}, RQ-Transformer \cite{rqtrans}, StyleGAN-XL \cite{sgxl}, ADMG \cite{admg}, ADMG-ADMU \cite{admg}
    \item FFHQ \cite{ffhq}: Efficient-vdVAE \cite{vae}, InsGen \cite{insgen}, LDM \cite{ldm}, StyleGAN-XL \cite{sgxl}, StyleGAN2-ada \cite{sg2}, StyleNAT \cite{stylenat}, StyleSwin \cite{styleswin}, Unleashing-transformers \cite{unleash}, Projected GAN \cite{progan}
\end{itemize}

\section{Parameter settings}
We examine optimal parameter settings for computing \com and \vul.
\cref{tab: params} shows the average \com and \vul of real and generated datasets with different parameter settings (i.e., $\epsilon$, $\alpha$, $K$, $J$) by using ConvNeXt as a feature model.
In \cref{tab: params}, \com of the generated dataset (by PFGM++) is smaller than that of the real dataset (CIFAR10) and \vul of the generated dataset is larger than that of the real dataset with parameter changes. 
The overall tendency of the \com and \vul is not affected by parameter changes. 

\begin{table}[h!]
\centering
\begin{tabular}{cc|cc|cc}
& & \multicolumn{2}{c|}{\com} & \multicolumn{2}{c}{\vul} \\
& & real & generated & real & generated \\
\midrule
\parbox[t]{2mm}{\multirow{3}{*}{\rotatebox[origin=c]{90}{{ $\epsilon$ \& $\alpha$}}}}&0.05 & 0.184 & 0.181 & 35.97 & 36.19 \\
&\textbf{0.01} & 0.099 & 0.098 & 14.57 & 15.24 \\
&0.005 & 0.080 & 0.076 & 7.77 & 7.95 \\
\multicolumn{2}{c|}{$K$\&$J$=5} & 0.098 & 0.069 & 7.24 & 7.45\\
\end{tabular}
\vspace{-.5em}
\caption{\textbf{\Com and \vul with various parameter settings.} Each cell denotes the average \com or \vul of the real or generated dataset. In the upper three rows, $K$ and $J$ are fixed as 10. In the last row, $\epsilon$ and $\alpha$ are 0.01.}
\label{tab: params}
\end{table}

\section{Two-tailed test for Tab. 1 and Tab. 2}
We report one-tailed tests in Tab. 1 and Tab. 2 of the main paper because we assume that \com and \vul of generated datasets are smaller than or larger than those of real datasets. For statistical clarity, we show the two-tailed test results in \cref{tab: two-tail} on the FFHQ dataset. 

\renewcommand{\arraystretch}{1}
\begin{table}[t]
\centering
\small
\begin{tabular}{cc|ccc}
\toprule
&&ViT&ConvNeXt&DINO-V2\\
\midrule
\multirow{3}{*}{\Com}& Reference&0.0643	&	0.0627	&	0.0311\\
&Generated&0.0638	&	0.0525	&	0.0302\\
&$p$-value&	0.4990	&	$<$0.0001$^*$	&	$<$0.0001$^*$	\\
\midrule
\multirow{3}{*}{\Vul}& Reference&18.30	&	14.57	&	12.90\\
&Generated&19.22	&	17.21	&	16.34\\
&$p$-value&$<$0.0001$^*$&$<$0.0001$^*$&$<$0.0001$^*$\\
\bottomrule
\end{tabular}
\caption{\textbf{\Com and \vul of FFHQ with two-tailed test.} We compare the average value of \com and \vul for various feature models, ViT-S \cite{vit}, ConvNeXt-tiny \cite{convnext}, and DINO-V2 \cite{dinov2}. `Reference' indicates the original dataset, FFHQ \cite{ffhq}. `Generated' denotes the \com and \vul obtained from datasets generated by InsGen \cite{insgen} trained with FFHQ. `$p$-value' denotes the $p$-value of the two-tailed $t$-test under the null hypothesis that \com of the generated dataset is equal to that of the reference dataset. The cases with statistical significance are marked with `$*$'.
}
\label{tab: two-tail}
\centering
\end{table}

\section{Linear regression on vulnerability}
In Sec 3.2, we explore the motivation of \vul by calculating the contributions of super-pixels of images to the changes caused by adversarial attacks.
We randomly select 3 to 6 super-pixels, add adversarial perturbations into them, and obtain the changes in the features due to the perturbations.
We repeat this process 20 times.
Then, we apply linear regression between the feature change and the set of binary variables indicating whether each super-pixel is attacked or not.
The linear regression is described as:
$\textrm{Y} = \textrm{V} \textrm{W} + b,$
where Y is a 20 (\# of trials)$\times$1 vector of the feature change, V is a 20 (\# of trials)$\times$20 (\# of super-pixels) matrix of variables that indicate whether each super-pixel is selected or not on each trial, W is a 20 (\# of super-pixels)$\times$1 vector of the linear regression coefficient, and $b$ is a 20 (\# of trials)$\times$1 vector of bias.
We consider the linear regression coefficient W as the contribution of each super-pixel to the feature changes, i.e., \vul.
If the coefficient is large, the corresponding super-pixel greatly contributes to the vulnerability.
On the other hand, if the coefficient is small, the corresponding super-pixel contributes less to the vulnerability.

\section{Results on various feature models}

\renewcommand{\arraystretch}{1}
\begin{table}[b]
\vspace{-1em}
\begin{minipage}{.5\linewidth}
\centering
\small
\begin{tabular}{cc|ccc}
\toprule
\multicolumn{2}{c}{\Com}&ResNet50&CLIP&DINO\\
\midrule
\multirow{3}{*}{CIFAR10}& Reference&0.1900	&	1.9246	&	0.0647\\
&Generated&0.1921	&	1.9234	&	0.0626\\
&$p$-value&-	&	0.0853	&$<$0.0001$^*$\\
\midrule
\multirow{3}{*}{ImageNet}& Reference&0.1170	&	1.9543	&	0.0326\\
&Generated&0.1190	&	1.9366	&	0.0331\\
&$p$-value&-&$<$0.0001$^*$&-\\
\midrule
\multirow{3}{*}{FFHQ}& Reference&0.1273	&	1.9899	&	0.0424\\
&Generated&0.1233	&	1.9893	&	0.0352\\
&$p$-value&$<$0.0001$^*$&0.1489&$<$0.0001$^*$\\
\bottomrule
\end{tabular}
\end{minipage} 
\begin{minipage}{.45\linewidth}
\centering
\small
\begin{tabular}{cc|ccc}
\toprule
\multicolumn{2}{c}{\Vul}&ResNet50&CLIP&DINO\\
\midrule
\multirow{3}{*}{CIFAR10}& Reference&44.59	&	6.67	&	37.40\\
&Generated&44.31	&	6.69	&	35.56\\
&$p$-value&-	&	$<0.05$$^*$	&	-\\
\midrule
\multirow{3}{*}{ImageNet}& Reference&32.18	&	4.54	&	9.27\\
&Generated&35.58	&	5.12	&	11.98\\
&$p$-value&$<$0.0001$^*$&$<$0.0001$^*$&$<$0.0001$^*$\\
\midrule
\multirow{3}{*}{FFHQ}& Reference&30.22	&	4.52	&	13.9\\
&Generated&30.85	&	4.39	&	12.57\\
&$p$-value&$<$0.0001$^*$&-&-\\
\bottomrule
\end{tabular}
\end{minipage}
\vspace{-.5em}
\caption{\textbf{\Com and \vul of various datasets.} We compare the average value of \vul for various feature models, ResNet50 \cite{resnet}, CLIP \cite{clip}, and DINO \cite{dino}. `Reference' indicates the original dataset, such as CIFAR10 \cite{cifar}, ImageNet \cite{imagenet}, and FFHQ \cite{ffhq}. `Generated' denotes datasets generated by PFGM++ \cite{pfgm++}, RQ Transformer \cite{rqtrans}, and InsGen \cite{insgen} trained with the respective reference datasets.
`$p$-value' denotes the $p$-value of the one-tailed $t$-test under the null hypothesis that \com or \vul of the generated dataset is equal to that of the reference dataset. The cases with statistical significance are marked with `$*$'. `-' means that the expectation is not met, i.e., \com (\vul) of the generated dataset is larger (smaller) than that of the reference dataset.
}
\label{tab: sup-vul}
\centering
\end{table}

We use six feature models, ResNet50 \cite{resnet}, ViT-S \cite{vit}, ConvNeXt-tiny \cite{convnext}, CLIP \cite{clip}, DINO \cite{dino}, and DINO-V2 \cite{dinov2}. Here, we present additional experimental results on these feature models, which are not included in the main paper.

\subsection{Complexity and vulnerability}
\label{subsec: sup-com-and-vul}
\cref{tab: sup-vul} indicates the average values of \com and \vul of the reference datasets and generated datasets when we use ResNet50, CLIP, and DINO as feature models.
In most cases, \com of the generated datasets is smaller than that of the reference datasets. \Vul of the generated datasets is larger than that of the reference datasets except for a few cases. These results are generally consistent with the results in the main paper (Tab. 1 and Tab. 2). However, in some cases using ResNet50 and DINO, the results are not aligned with our assumption, implying that they are less preferable as the feature model of our method.

\begin{figure*}[h]
\centering
    \includegraphics[width=0.19\columnwidth]{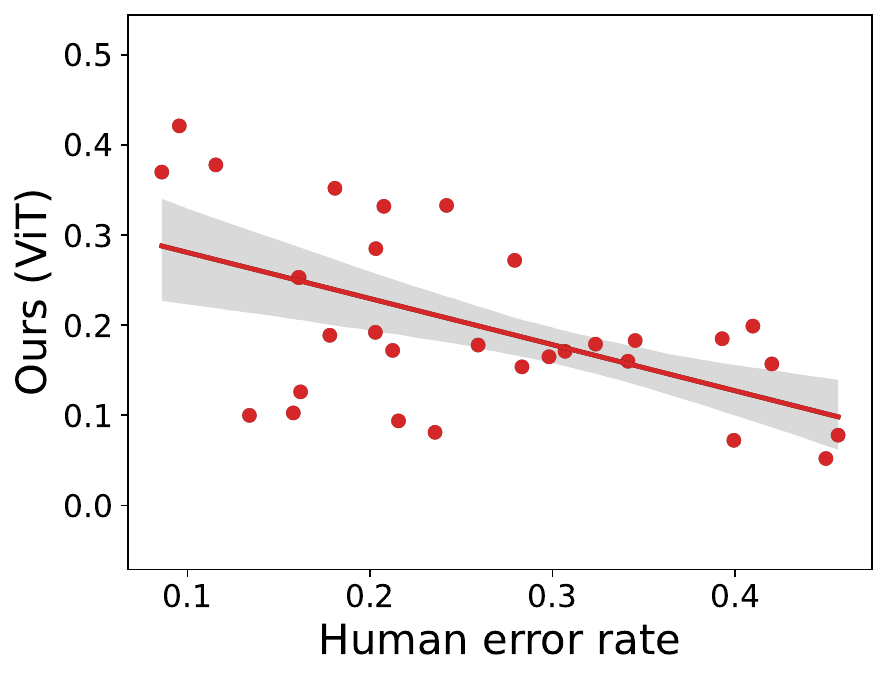}
    \includegraphics[width=0.19\columnwidth]{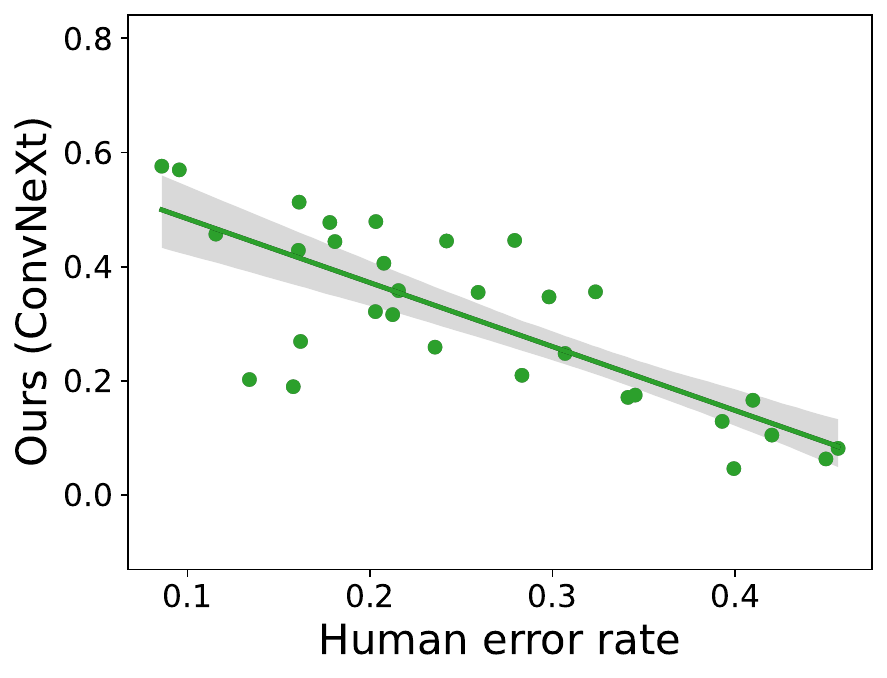}
    \includegraphics[width=0.19\columnwidth]{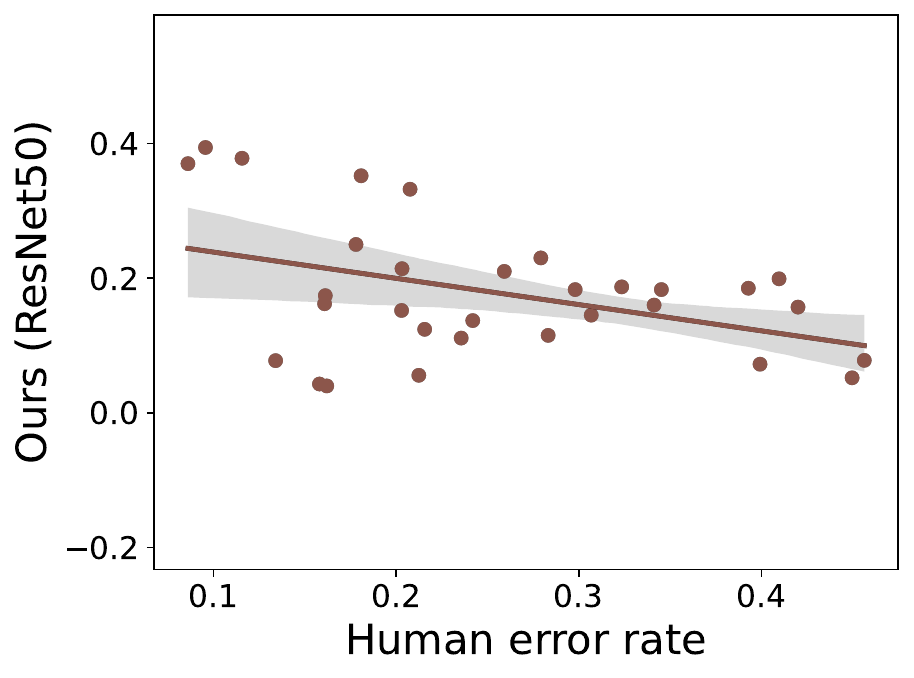}
    \includegraphics[width=0.19\columnwidth]{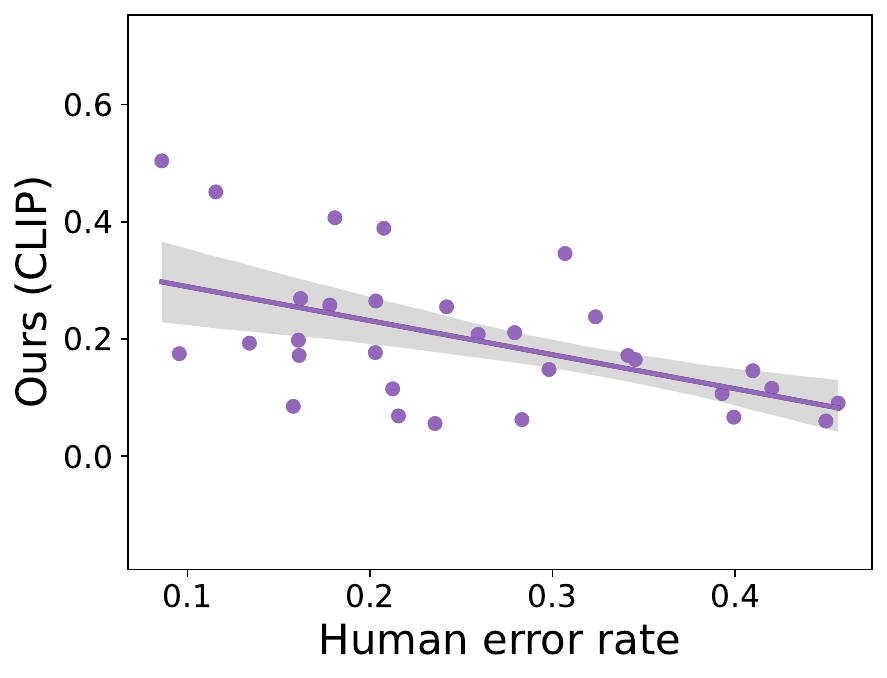}
    \includegraphics[width=0.19\columnwidth]{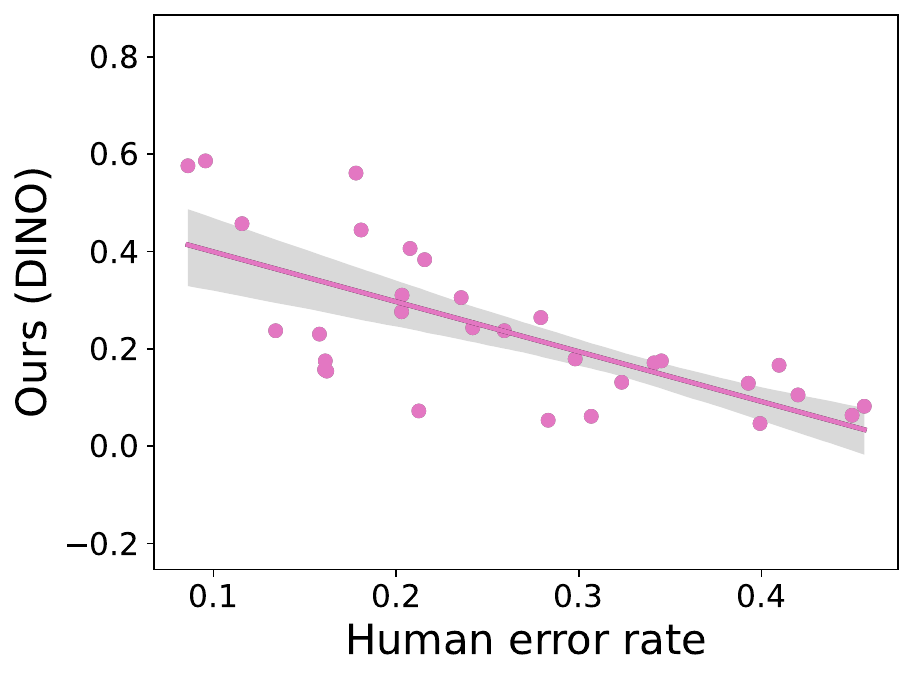}
\caption{\textbf{Performances of our method using various models for overall datasets.} Each dot represents a distinct dataset generated by a generative model. A high human error rate indicates a high-quality dataset, while a high AS score means a low-quality dataset.}
\label{fig:all-all-anomaly-sup}
\vspace{-.5em}
\end{figure*}

\begin{figure*}[h]
\centering
\begin{subfigure}[b]{\textwidth}
    \centering
    \includegraphics[width=0.24\columnwidth]{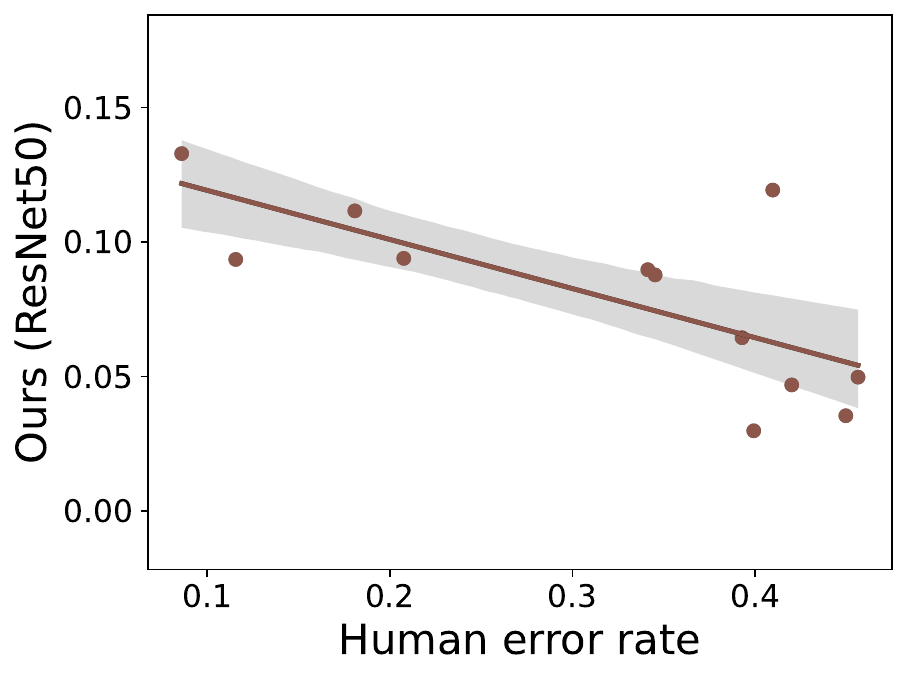}
    \includegraphics[width=0.24\columnwidth]{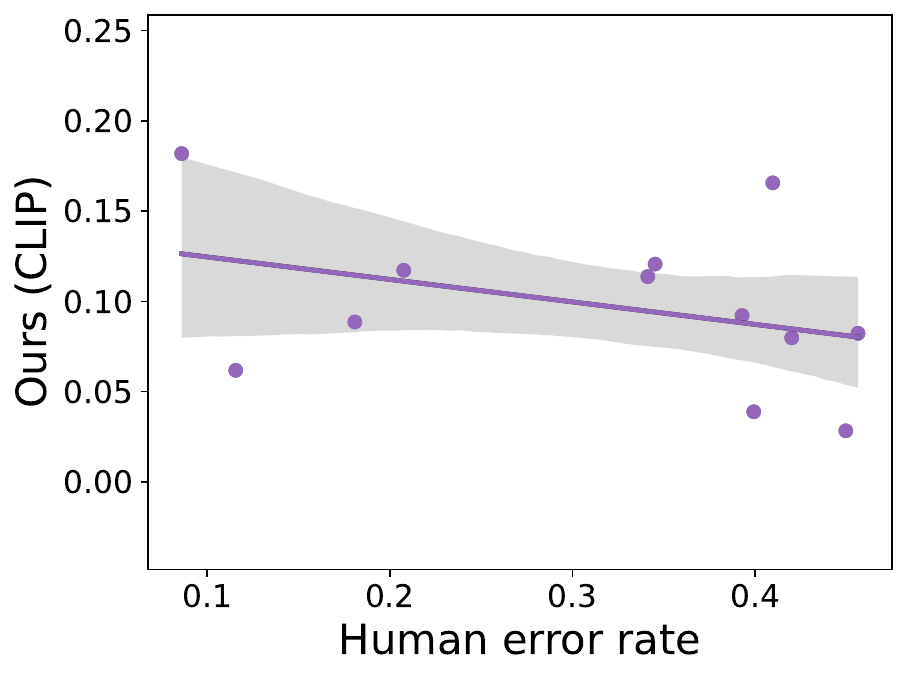}
    \includegraphics[width=0.24\columnwidth]{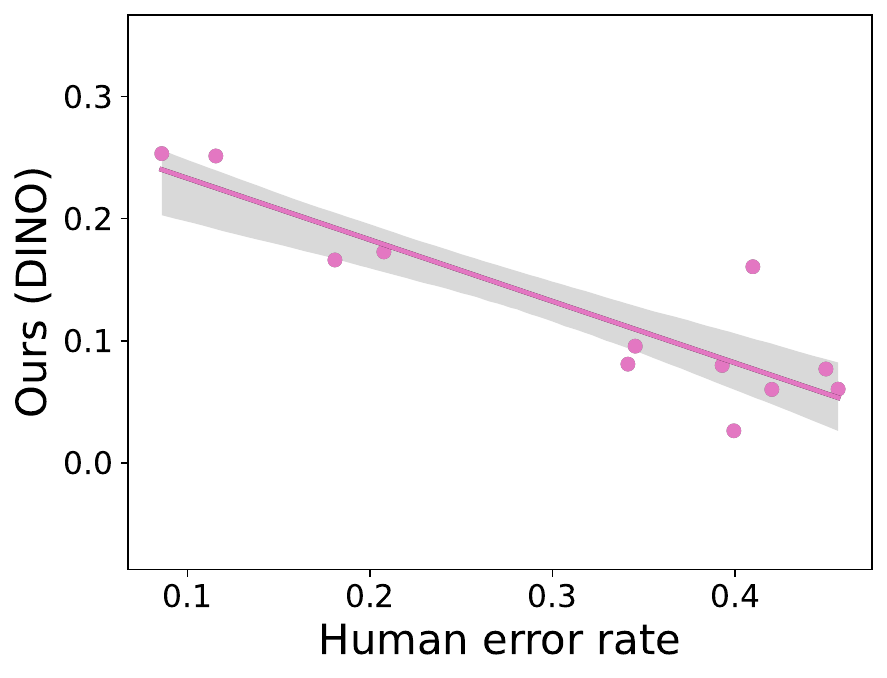}
    \includegraphics[width=0.24\columnwidth]{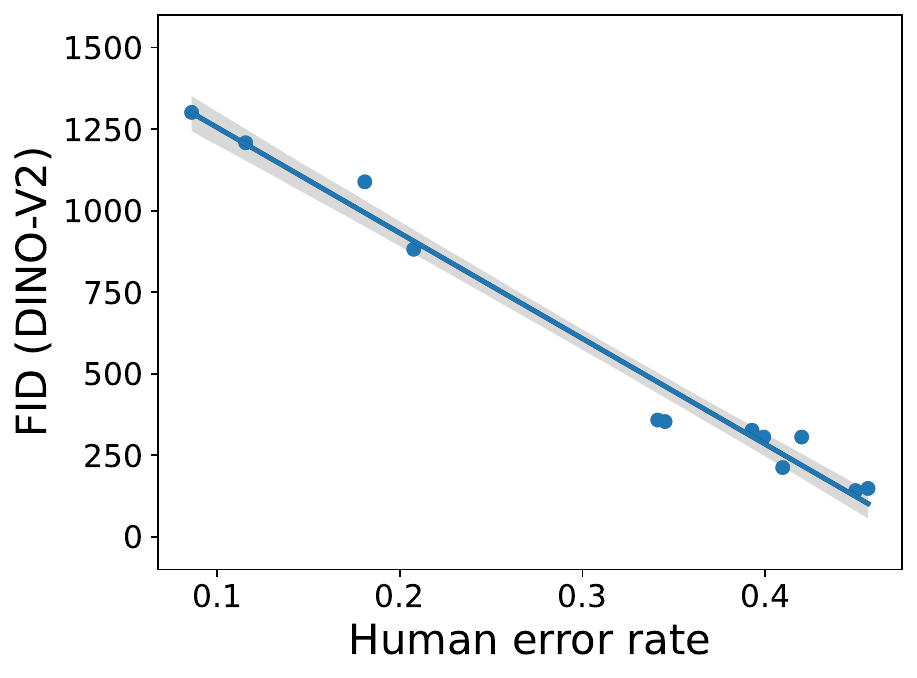}
    \label{fig:}
    \caption{CIFAR10}
\end{subfigure}
\begin{subfigure}[b]{\textwidth}
\centering
    \includegraphics[width=0.24\columnwidth]{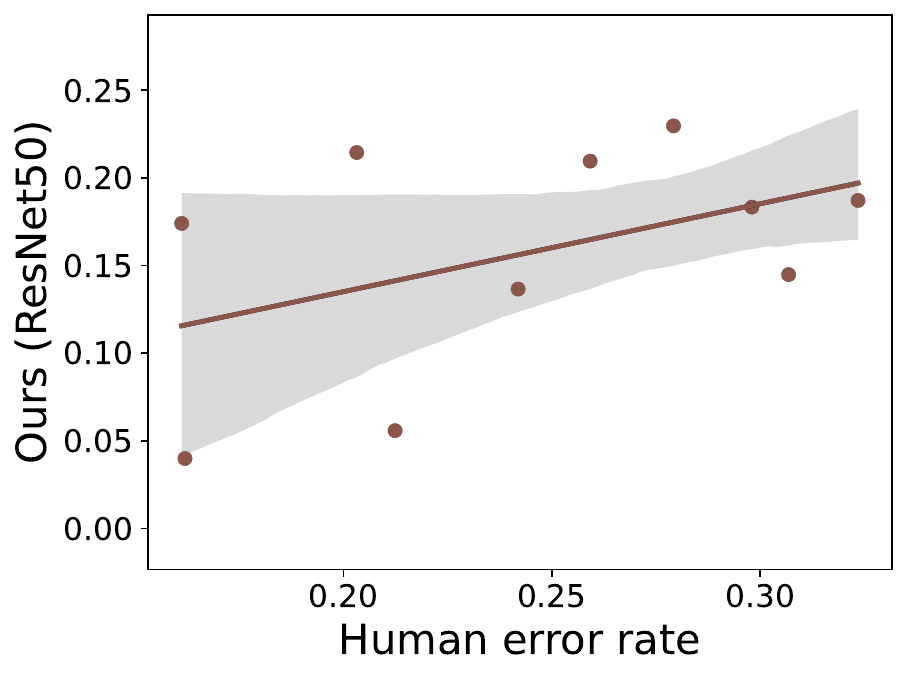}
    \includegraphics[width=0.24\columnwidth]{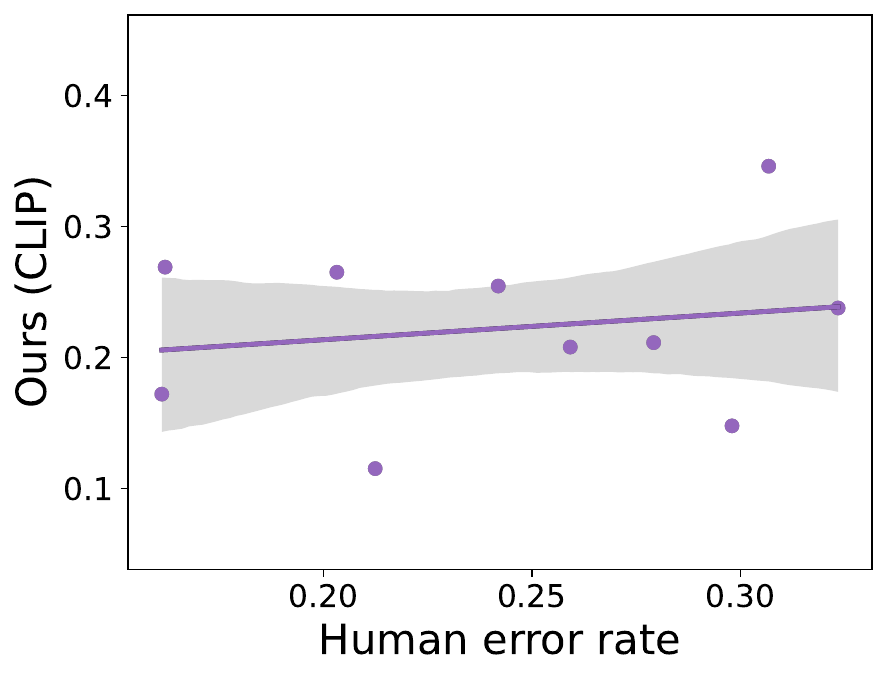}
\includegraphics[width=0.24\columnwidth]{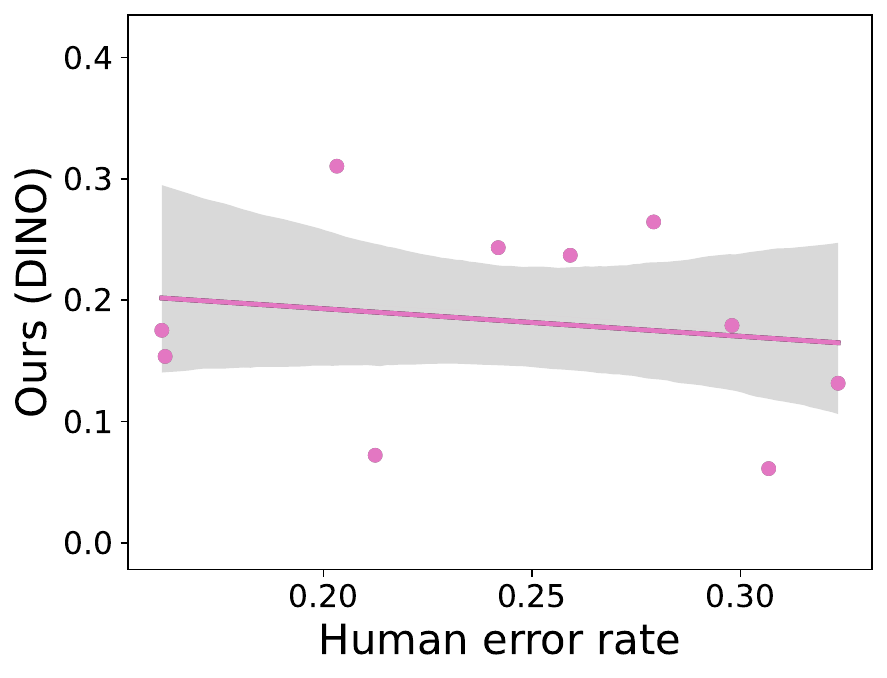}
\includegraphics[width=0.24\columnwidth]{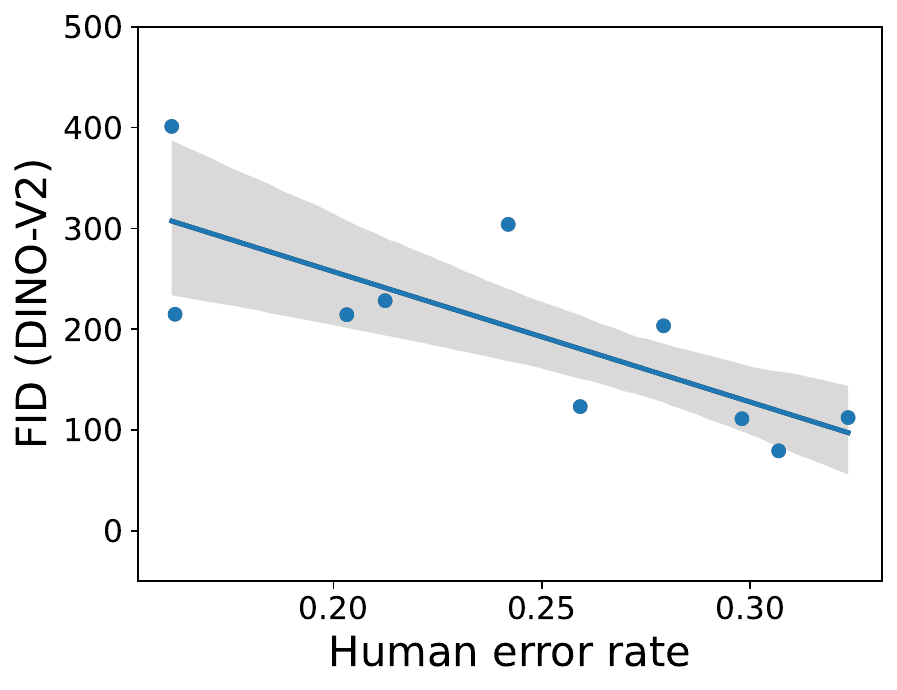}
    \label{fig:}
    \caption{ImageNet}
\end{subfigure}

\begin{subfigure}[b]{\textwidth}
    \centering
\includegraphics[width=0.24\columnwidth]{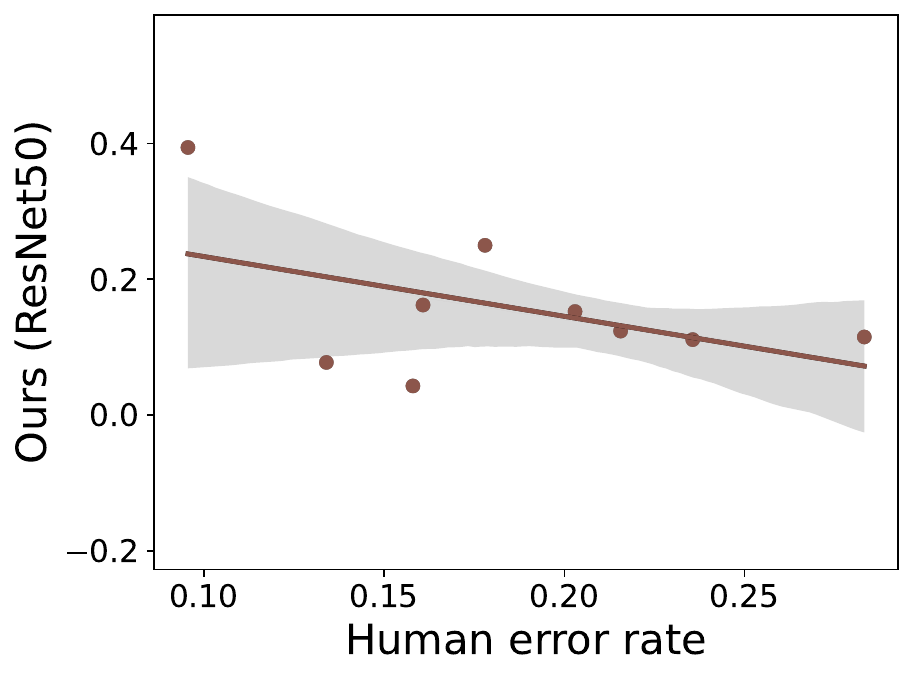}
\includegraphics[width=0.24\columnwidth]{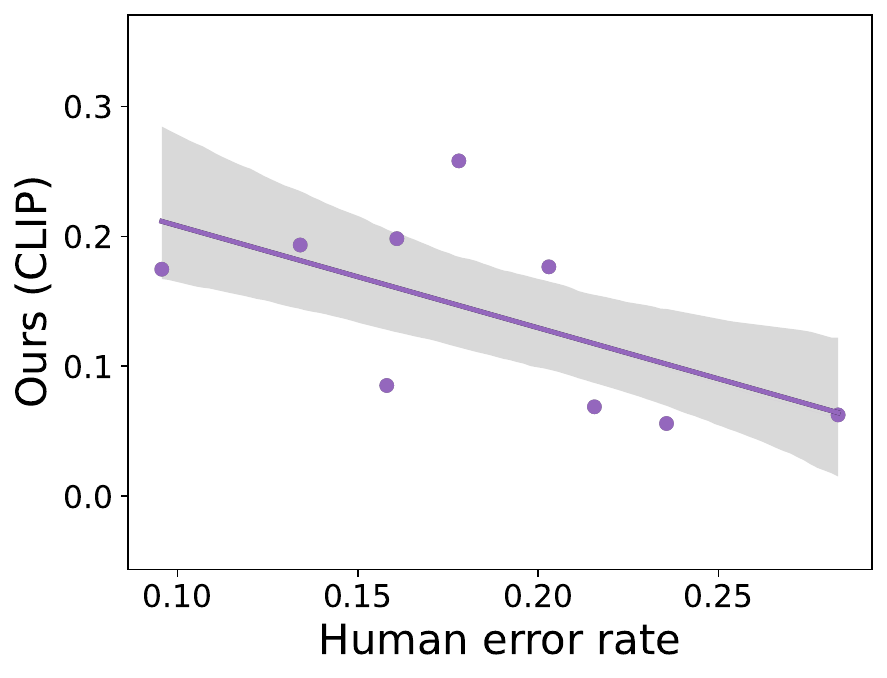}
\includegraphics[width=0.24\columnwidth]{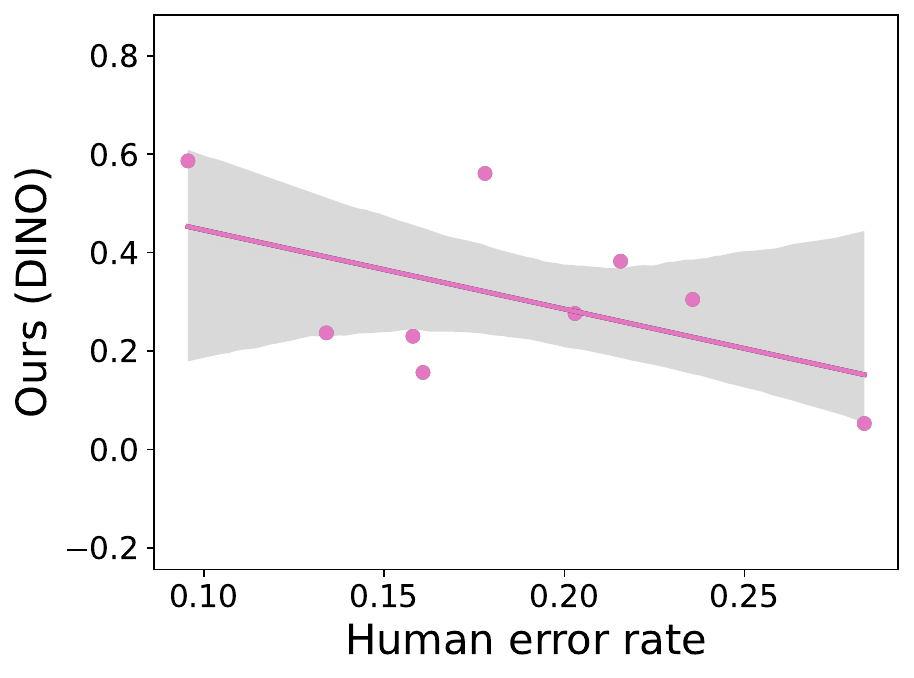}
\includegraphics[width=0.24\columnwidth]{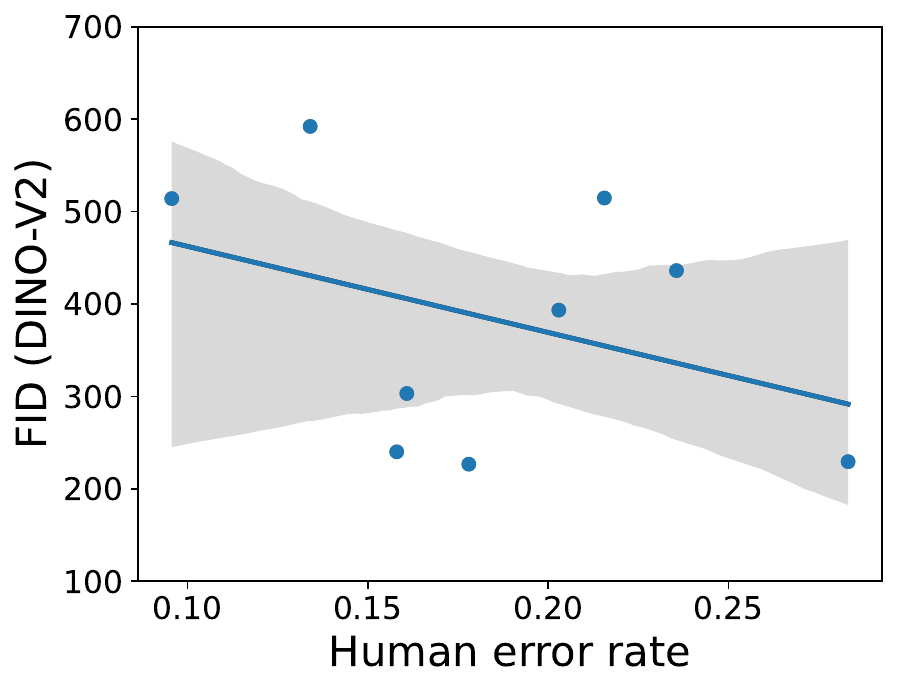}
    \label{fig:}
    \caption{FFHQ}
\end{subfigure}
\caption{\textbf{Overall results of evaluating generative models on various datasets.} Each dot represents a distinct dataset generated by a generative model. A high human error rate indicates a high-quality dataset, while a high AS score means a low-quality dataset. The first three columns show AS with different feature models: ResNet50, CLIP, and DINO, respectively. The last column is the result of FID~\cite{FID} with the DINO-V2 model.}
\label{fig:all-anomaly-sup}
\vspace{-.5em}
\end{figure*}

\subsection{Anomaly score}

\cref{fig:all-all-anomaly-sup} indicates evaluation results of all generative models targeting all image datasets (CIFAR10, ImageNet, and FFHQ) using the proposed AS with various feature models except for DINO-V2. The results with DINO-V2 are shown in Fig. 8 of the main paper.

\cref{fig:all-anomaly-sup} shows evaluation results of various generative models using AS with ResNet50, CLIP, and DINO as feature models and FID with DINO-V2. In the case of CIFAR10 and FFHQ, AS correlates well with human perception (-0.72, -0.36, and -0.89 pearson correlation coefficients (PCCs) on CIFAR10, and -0.47, -0.60, and -0.51 PCCs on FFHQ, respectively). On the other hand, AS with ResNet50, CLIP, and DINO shows low correlations on generated datasets for ImageNet (0.45, 0.17, and -0.16 PCCs, respectively).
Due to the weak alignment between the characteristics of the representation space of ResNet50, CLIP, and DINO and our assumptions (\cref{subsec: sup-com-and-vul}), the performance of the anomaly score using them is lower than that using ViT-S, ConvNeXt-tiny, and DINO-V2.

\section{Comparison with Inception-V3}

\begin{figure}[t]
    \centering
    \begin{subfigure}[b]{0.3\linewidth}
        \includegraphics[width=\textwidth]{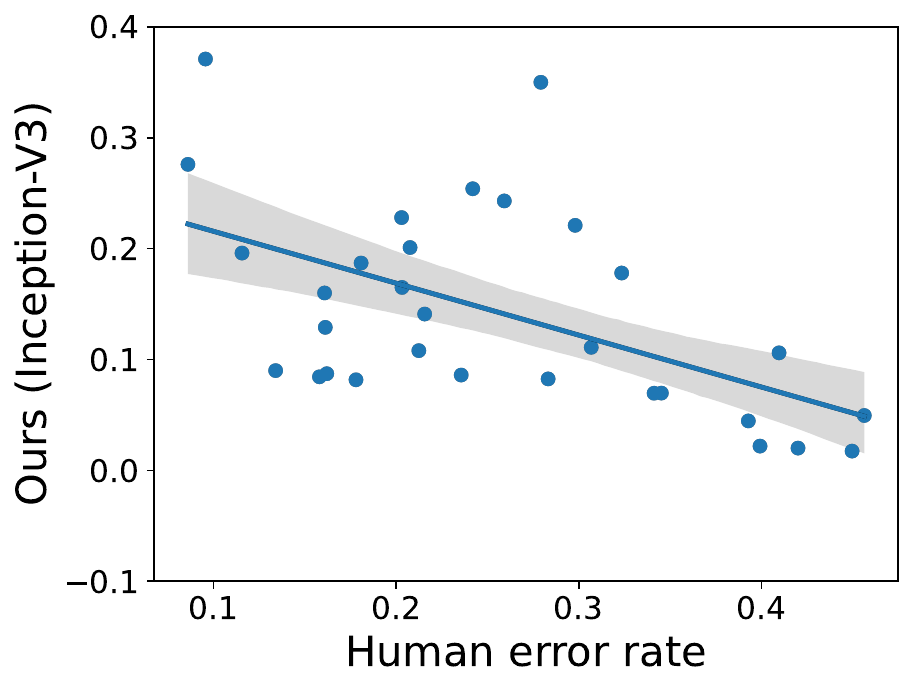}
        \caption{Ours (Inception-V3)}
        \label{fig:perform-a}
    \end{subfigure}
    \begin{subfigure}[b]{0.3\linewidth}
        \includegraphics[width=\textwidth]{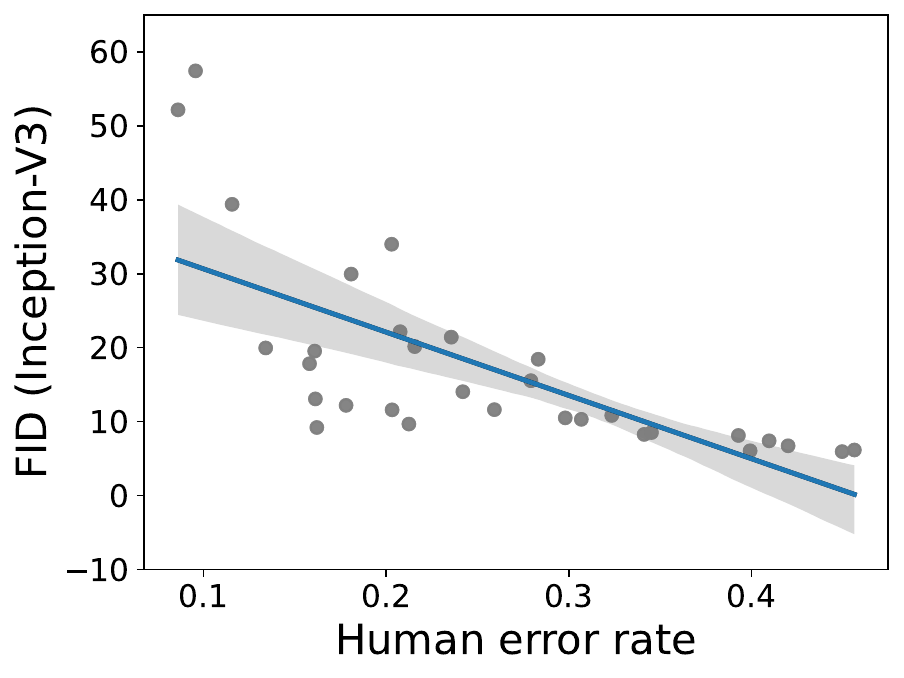}
        \caption{FID (Inception-V3)}
        \label{fig:perform-c}
    \end{subfigure}
    \vspace{-.5em}
    \caption{\textbf{Performances of our method and FID using Inception-V3.} We evaluate various generative models by the proposed method and FID with Inception-V3. Each dot represents a distinct dataset generated by a generative model. A high human error rate indicates a high-quality dataset, while a high AS score means a low-quality dataset.}
    \label{fig:perform-inc}
\end{figure}

In Sec. 4 of the main paper, we mainly use DINO-V2 as a feature model for FID since it shows high performance in \cite{dgm-eval}.
\cref{fig:perform-inc} shows the evaluation results using AS and FID with Inception-V3 \cite{Inception} as a feature model. Experimental settings for evaluation including used generative models and parameter settings are the same as those of Fig. 8 of the main paper. The PCC of ours is -0.54, which has a comparatively weaker correlation than one of our methods using DINO-V2 (-0.81). On the other hand, FID using Inception-V3 shows a comparatively stronger correlation (PCC=-0.71) compared to FID using DINO-V2 (PCC=-0.54). However, FID using Inception-V3 provides poor evaluation performance on generative models targeting ImageNet \cite{dgm-eval}. Thus, in the main paper, we mainly compare our method with FID using DINO-V2.

\section{Transformation of anomaly score}
\begin{figure}[h]
    \centering
        \includegraphics[width=0.3\textwidth]{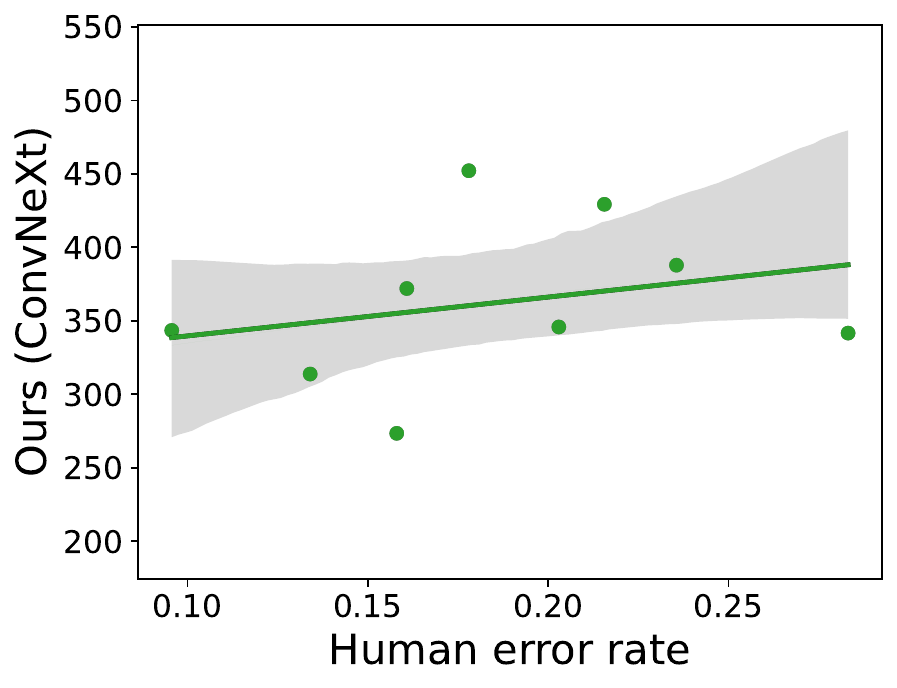}
    \vspace{-.5em}
    \caption{\textbf{Performance of the average of AS-i.} We evaluate generative models targeting FFHQ by the average of AS-i using ConvNeXt as a feature model. Each dot represents a distinct generated dataset. A high human error rate indicates a high-quality dataset, while a high average of AS-i means a low-quality dataset.}
    \label{fig:perform-asi}
\end{figure}

We define anomaly score by comparing the distributions of \com and \vul. Here, we provide the additional experimental results when we evaluate generative models using the average of individual AS-i.
For evaluating each generated dataset targeting FFHQ utilizing ConvNeXt as a feature model, we first compute the individual score, AS-i, of each image and then take the average across the images in the dataset. 
As shown in \cref{fig:perform-asi}, the average of AS-i does not work well in evaluating generative models.
This seems to be because numerical differences in AS-i is limited to capture distributional differences between real and generated datasets.

\section{Images for subjective test}

\begin{figure}[t]
    \includegraphics[width=0.9\columnwidth]{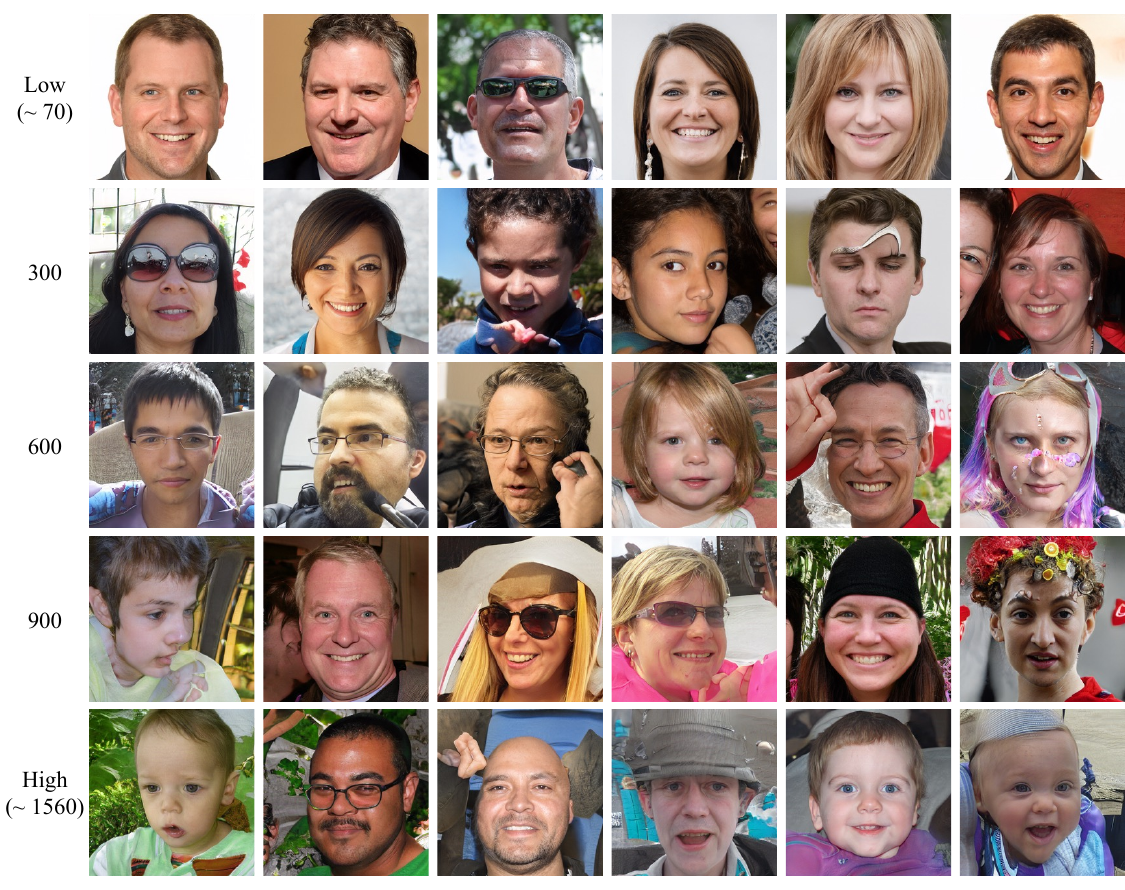}
\caption{\textbf{Examples having various levels of AS-i.}}
    \label{fig:sup-ex-level}
\end{figure}

In Sec. 5 of the main paper, we evaluate our anomaly score for individual images, AS-i, by conducting the subjective test with 20 images for each AS-i level. \cref{fig:sup-ex-level} shows example images according to each AS-i level. If an image has a low AS-i level, the image looks natural and clear, like real images.
Images with higher AS-i levels contain more unnatural components, such as abnormal patterns in faces and backgrounds.
\cref{fig:sup-ex-level} shows that the severity of the unnatural pattern in the image increases as the AS-i level increases.

\end{document}

%% file: preamble.tex
\usepackage[dvipsnames]{xcolor}


%% file: sec/0_abstract.tex
\begin{abstract}
With the advancement of generative models, the assessment of generated images becomes increasingly more important. Previous methods measure distances between features of reference and generated images from trained vision models. In this paper, we conduct an extensive investigation into the relationship between the representation space and input space around generated images. We first propose two measures related to the presence of unnatural elements within images: complexity, which indicates how non-linear the representation space is, and vulnerability, which is related to how easily the extracted feature changes by adversarial input changes. Based on these, we introduce a new metric to evaluating image-generative models called anomaly score (AS). Moreover, we propose AS-i (anomaly score for individual images) that can effectively evaluate generated images individually. Experimental results demonstrate the validity of the proposed approach.

\end{abstract}

%% file: sec/1_intro.tex
\section{Introduction}
\label{sec:intro}

\newcommand{\Frechet}[0]{Fr\'echet~}

The advancement of deep learning has significantly fostered the development of generative AI, particularly in the domain of image generation. Initially, the focus was primarily on generative adversarial networks (GANs)~\cite{goodfellow2014generative,ffhq,sg2,sgxl}, which employed a generator and a discriminator. Recently, various generative models have been suggested, including autoencoder-based models~\cite{vae,evdvae} and diffusion-based models~\cite{diffusion,ddpm,adm,ldm}. Simultaneously, evaluating the performance of generative models has become more critical.

\begin{figure}[t]
    \centering
    \includegraphics[width=0.95\linewidth]{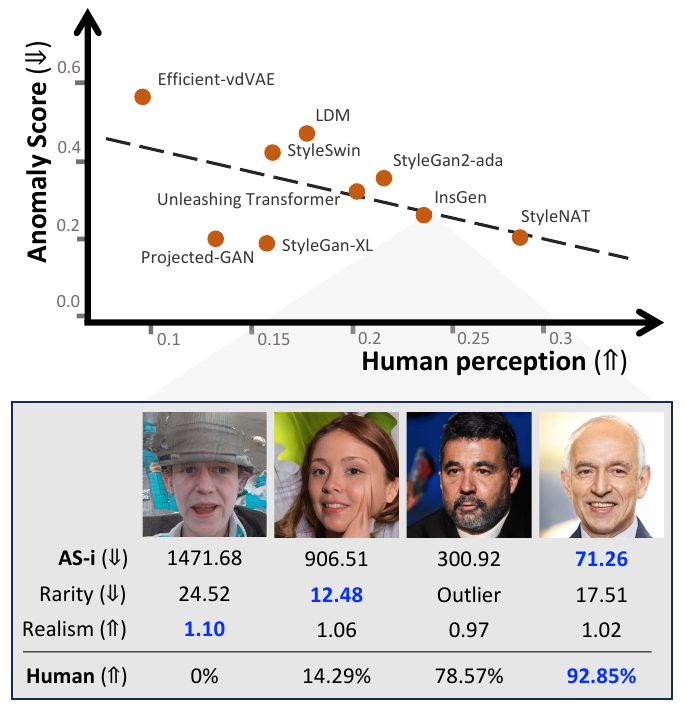}
    \caption{\textbf{Proposed AS for evaluating generative models and AS-i for individual images.} The graph on the top shows the proposed AS aligns well with the human perception of evaluating various generative models trained on the FFHQ dataset. On the bottom, several generated images are shown with our AS-i score, rarity score \cite{rarity}, realism score \cite{IPR}, and human evaluation.
    A value of the human evaluation indicates the proportion of participants who assess that the image is a natural image in our subjective test.
    The best score for each metric, indicating an image to be the most natural image is highlighted in blue.
    In terms of naturalness, AS-i shows the best alignment with human evaluation. 
    On the other hand, the rarity score prefers the second image, which is unnatural, as the most common in real images. The realism score also overestimates the leftmost image to be the most realistic.
    }
    \label{fig:teaser}
    \vspace{-1.5em}
\end{figure}

The performance of generative models can be evaluated in various ways~\cite{ganeval}. 
Assessing how similar generated images are to real images (i.e., fidelity) is one of the most important and challenging aspects of evaluating the performance.
An accurate approach is conducting subjective tests, where human subjects are asked to judge the naturalness of generated images, but this is resource-intensive.
To efficiently evaluate generative models in various aspects, several objective metrics have been proposed~\cite{IS,FID,KID,PR,IPR,rarity,eval1,eval2,eval3,infinity}.
Most existing metrics involve comparing sample statistics between the sets of real and generated images after extracting features from pre-trained vision models.
For instance, the \Frechet inception distance (FID) utilizes the Inception model~\cite{Inception} to extract Inception features and measures the 2-Wassertein distance of the Inception features between the real and generated datasets.

However, it is argued that such metrics are often misaligned with the human judgment on naturalness.
In~\cite{role}, it is shown that FID has a null space where the score is unaffected by the change of naturalness of generated images.
Furthermore, \cite{dgm-eval} shows that the score often focuses on image parts unrelated to the naturalness. 
To sum up, existing metrics are subject to inconsistency to a certain extent in assessing the naturalness of generated images.

We argue that relying only on feature distances is insufficient to assess the naturalness of generated images.
Consider a pair of real images having a certain distance in a representation space (i.e., feature space).
We can modify one of the images (e.g., by adding noise) so that the modified one is at the same certain distance from the original image.
In this case, while the feature distance between the two real images is the same as the distance between the chosen real image and its modified version, the modified image has significantly different contents in terms of naturalness.
This also applies to generated images, having certain distances to real images may not be able to accurately represent whether the generated images are natural or not.

In order to address this limitation, in this paper, we propose two novel metrics: anomaly score (AS) for evaluating generative models and anomaly score for individual images (AS-i) for evaluating individual generated images.
Instead of simply measuring distances between features, AS and AS-i capture the relationship between the input space and the representation space based on two new perspectives, \com and \vul.
We demonstrate that both metrics show significant correlations with the human-perceived naturalness of generated images (\cref{fig:teaser}).

We define \com as the amount of variations in the direction of feature changes with respect to linear input changes.
A trained neural network model typically implements a non-linear function, and the degree of non-linearity (which we refer to as \com) depends on the feature location in the representation space \cite{juyeop}.
We observe that \com tends to become larger for real images compared to unnatural generated images.

\Vul reflects how easily the extracted feature of an image is changed due to adversarial input changes.
We apply the concept of adversarial attacks \cite{fgsm, pgd, autoattack}, which point out weaknesses of deep learning models and are also utilized for tasks capturing characteristics of images, such as out-of-distribution detection \cite{odin}.
We observe that \vul tends to be smaller for real images compared to generated unnatural images.
Our contributions are summarized as follows.
\begin{enumerate}
    \item 
    We introduce \com and \vul, to examine the characteristics of the representation space.
    \Com measures how non-linear the representation space around an image is with respect to the linear input changes. And, \vul captures how easily an extracted feature is changed by adversarial input changes. We demonstrate that \com and \vul of generated images are significantly different compared to those of real images.
    \item 
    We propose a novel metric called anomaly score (AS) to evaluate generative models in terms of naturalness based on \com and \vul. AS is the difference of joint distributions of \com and \vul between the sets of reference real images and generated images, which is quantified by 2D Kolmogorov-Smirnov (KS) statistics.
    Our method aligns better human judgment about the unnaturalness of generated images compared to the existing method.  
    \item 
     We suggest the anomaly score for individual images (AS-i) to assess generated images individually. By conducting subjective tests, we demonstrate that AS-i outperforms existing methods for image evaluation.

\end{enumerate}

%% file: sec/2_relatedwork.tex
\section{Related works}
\label{sec:related-works}

\subsection{Generative models}
Generative models have garnered significant attention for their ability to generate realistic data samples.
GANs~\cite{goodfellow2014generative,ffhq,sg2,sgxl} leverage a game-theoretic approach, employing a generator and discriminator in a competitive setting.
Variational autoencoders (VAEs)~\cite{vae,evdvae}, on the other hand, model the distribution of the training data with a likelihood function, learning latent variable representations to generate data that closely matches the observed distribution.
Recently, diffusion models~\cite{diffusion,ddpm,adm,ldm} have emerged as a powerful approach in generating high-quality images and capturing complex data distributions.

\subsection{Evaluation of generative models}
Evaluation of generative models mostly involves comparing sample statistics between the generated data and the real target data.
Existing metrics can be categorized into three groups based on the way of evaluation: summarizing overall performance of generative models in a single score~\cite{IS,FID,KID}, evaluating different aspects of performance (e.g., fidelity and diversity) of models separately~\cite{PR,IPR,DC}, and assessing individual generated images~\cite{IPR,rarity}.

In the first category, FID~\cite{FID} is one of the most widely used metrics, which measures how well a generative model can reproduce the target data distribution.
It employs the trained Inception model~\cite{Inception} to extract features from the generated and the real images and measures the 2-Wasserstein distance between the two feature distributions.
However, it often fails to model the density of the feature distributions~\cite{trend} and to align with human perception~\cite{role}.

For the second category, Precision and Recall~\cite{PR,IPR} measure fidelity and diversity of samples from generative models, respectively, by extending the classical precision and recall metrics for machine learning.
They construct manifolds of the real samples and the generated samples in a certain representation space, then count the proportions of generated and real samples that belong to the real and generated manifolds, respectively.
While such twofold metrics are effective for a diagnostic purpose, they are often vulnerable to outliers~\cite{DC} and suffer from high computational costs for measuring pairwise distances between samples.

Few studies have addressed evaluation of individual samples~\cite{IPR,rarity}.
The realism score~\cite{IPR} examines the relationship between a generated image and the real manifold.
Still, it is based on the manifold that is vulnerable to outliers, thus may become deviated from human evaluation (as will be shown in our experiments).
The rarity score~\cite{rarity} focuses on how rare or uncommon a generated image is in order to consider the performance of generative models in terms of creativity. Thus, the rarity score does not provide accurate information about the naturalness of an image.




%% file: sec/3_why.tex
\section{Analyzing representation space around generated images}
\label{sec:why}

In this section, we demonstrate that the representation space, which is the space of features extracted by trained models for vision tasks, around generated images exhibits distinct properties in comparison to that around real images.

\begin{figure}
  \begin{center}
    \includegraphics[width=0.6\linewidth]{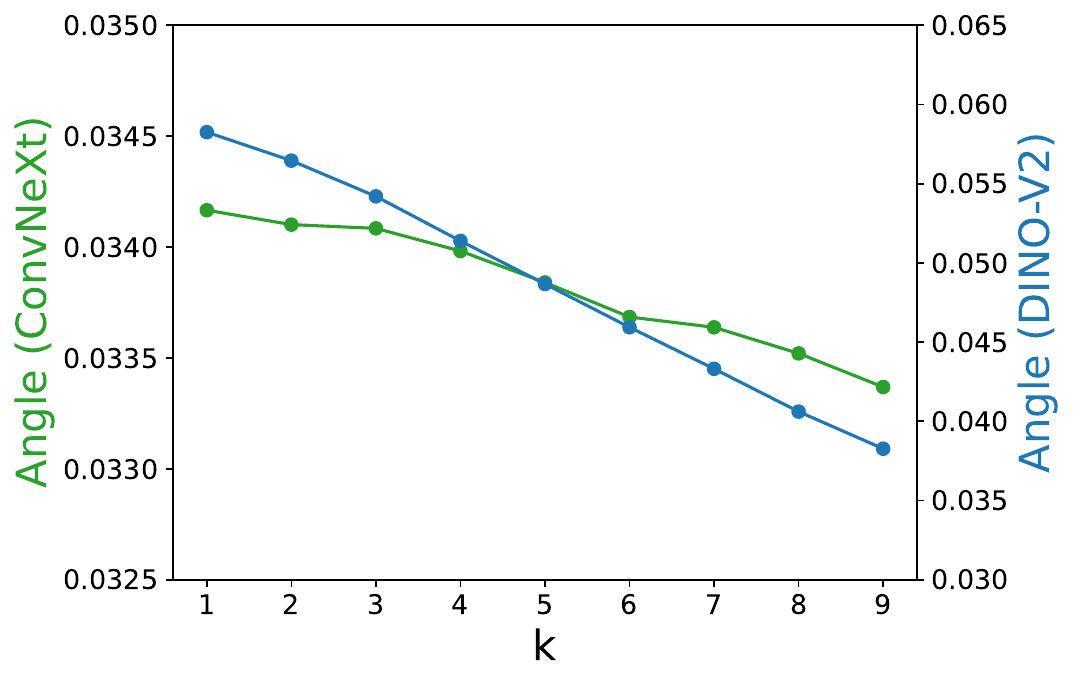}
  \vspace{-.5em}
  \caption{\textbf{Tendency of linearity around real data.} We compute the change in the linearity of representation spaces developed by ConvNeXt-tiny and DINO-V2 when random noise is added to real images from the ImageNet dataset.}
  \label{fig:k_angle}
  \vspace{-2em}
  \end{center}
\end{figure}

\subsection{Complexity}
\label{sec:complexity}

{\noindent \bf Motivation.}
When an input of a trained model changes linearly, it is expected that its feature representation (i.e., output of the model before the softmax function) does not change linearly but instead exhibits curvature due to the non-linearity inherent in deep neural networks.
In \cite{juyeop}, it is observed that the regions around the features of training images in the representation space appear curved (i.e., complex) after training, i.e., a linear movement in the input space yields a curved trajectory in the representation space;
on the other hand, the representation space near modified images with random noise is less complex.

To verify this, we compute the changes 
of the average angle between $M(\mathrm{x}^{k})-M(\mathrm{x}^{k-1})$ and $M(\mathrm{x}^{k+1})-M(\mathrm{x}^{k})$ with varying $k$, 
where $M(\cdot)$ indicates the model used for extracting features, which is referred to as feature model for simplicity, and $\mathrm{x}^{k}$ indicates a changed input $\mathrm{x}$ by adding a constant noise $k$ times. 
In \cref{fig:k_angle}, we can observe that the angle decreases as $k$ increases. 
This phenomenon suggests that the regions far from training images in the representation space are less complex when compared to those around the training images themselves. In a similar context, we hypothesize that the representation space around generated images is less complex than that surrounding reference real images.

{\noindent \bf Definition.}
To assess the \com of the representation space around an image, we gradually add random noises to the image and quantify the angular variations in the corresponding feature movements.
Let $\mathrm{x}$ and $\mathrm{N}$ denote an image and a Gaussian random noise vector having a unit length, respectively.
$\mathrm{x}$ is corrupted gradually by $\epsilon \mathrm{N}$, i.e., the noised image at step $k$ is computed as $\mathrm{x}^k = \mathrm{x}^0+k\epsilon \mathrm{N}$, where $\epsilon$ is the parameter controlling the magnitude of noise and $\mathrm{x}^0=\mathrm{x}$. 
Then, we calculate the angle between the changes of the output feature within two successive steps (i.e., $k-1$ and $k$).
For instance, when the features for $\mathrm{x}^{k-2}$, $\mathrm{x}^{k-1}$, and $\mathrm{x}^{k}$ are on a straight line, the angle is zero.
We compute the \com $\mathrm{C}(\cdot)$ by averaging the angles across the changes over multiple steps, which is formulated as follows.
\vspace{-.5em}
\begin{equation}
\tiny
    \mathrm{C}(\mathrm{x}) = \frac{1}{K-1}\sum_{k=1}^{K-1} \Big (  \cos^{-1}{\frac{(M(\mathrm{x}^{k})-M(\mathrm{x}^{k-1})) \cdot (M(\mathrm{x}^{k+1})-M(\mathrm{x}^{k}))}{||M(\mathrm{x}^{k})-M(\mathrm{x}^{k-1})||~||M(\mathrm{x}^{k+1})-M(\mathrm{x}^{k})||}} \Big ),
\label{eq:angle}
\end{equation}
where $M(\cdot)$ denotes the feature model and $K$ represents the total number of steps of adding random noise. 
Note that the feature model can be chosen among various models trained for vision tasks, including ImageNet classification models and models trained by self-supervised learning methods.

{\noindent \bf Experimental setup.}
We conduct an experiment to compare the \com defined above for real images and generated images.
We use six pre-trained models for the feature model ($M(\cdot)$): three supervised ImageNet classification models, ResNet50 \cite{resnet}, ViT-S \cite{vit}, and ConvNeXt-tiny \cite{convnext}, and three self-supervised models, DINO \cite{dino}, DINO-V2 \cite{dinov2}, and CLIP \cite{clip}.
We utilize three reference datasets, CIFAR10 \cite{cifar}, ImageNet \cite{imagenet}, and FFHQ \cite{ffhq}.
We employ generated datasets produced by PFGM++ \cite{pfgm++} trained with CIFAR10, RQ Transformer \cite{rqtrans} trained with ImageNet, and InsGen \cite{insgen} and StyleNAT \cite{stylenat} trained with FFHQ from dgm-eval \cite{dgm-eval}. And we utilize 10000 generated images from each dataset. 
We set $\epsilon=0.01$ and $K=10$.

\renewcommand{\arraystretch}{1}
\begin{table}[t]
\centering
\small
\begin{tabular}{cc|ccc}
\toprule
&&ViT&ConvNeXt&DINO-V2\\
\midrule
\multirow{3}{*}{CIFAR10}& Reference&0.1046	&	0.0986	&	0.0578\\
&Generated&0.1018	&	0.0975	&	0.0573\\
&$p$-value&	$<$0.0001$^*$	&	$<$0.0001$^*$	&	$<$0.0005$^*$	\\
\midrule
\multirow{3}{*}{ImageNet}& Reference&0.0519	&	0.0485	&	0.0337\\
&Generated&0.0410	&	0.0287	&	0.0102\\
&$p$-value&	$<$0.01$^*$	&	$<$0.0001$^*$	&	$<$0.0001$^*$	\\
\midrule
\multirow{3}{*}{FFHQ}& Reference&0.0643	&	0.0627	&	0.0311\\
&Generated&0.0638	&	0.0525	&	0.0302\\
&$p$-value&	0.2495	&	$<$0.0001$^*$	&	$<$0.0001$^*$	\\
\bottomrule
\end{tabular}
\caption{\textbf{\Com of various datasets.} We compare the average value of \com for various feature models, ViT-S \cite{vit}, ConvNeXt-tiny \cite{convnext}, and DINO-V2 \cite{dinov2}. `Reference' indicates the original dataset, such as CIFAR10 \cite{cifar}, ImageNet \cite{imagenet}, and FFHQ \cite{ffhq}. `Generated' denotes the \com obtained from datasets generated by PFGM++ \cite{pfgm++}, RQ Transformer \cite{rqtrans}, and InsGen \cite{insgen} trained with the respective reference datasets. `$p$-value' denotes the $p$-value of the one-tailed $t$-test under the null hypothesis that \com of the generated dataset is equal to that of the reference dataset. The cases with statistical significance are marked with `$*$'.
}
\label{tab: angle}
\centering
\end{table}

\begin{figure}
\centering
\begin{subfigure}[b]{0.32\columnwidth}
    \centering    \includegraphics[width=\columnwidth]{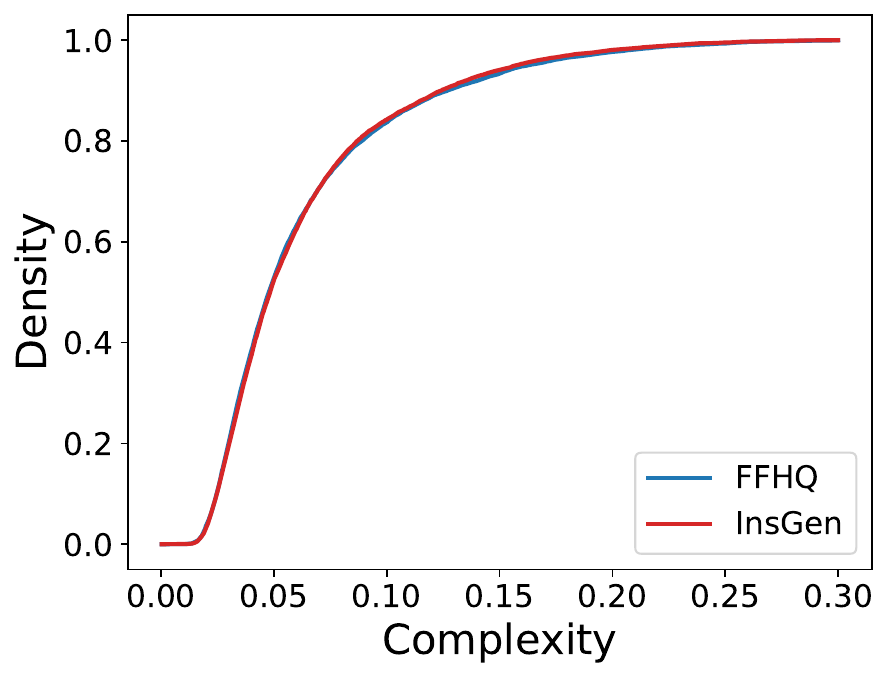}
\end{subfigure}
\begin{subfigure}[b]{0.32\columnwidth}
    \centering
    \includegraphics[width=\columnwidth]{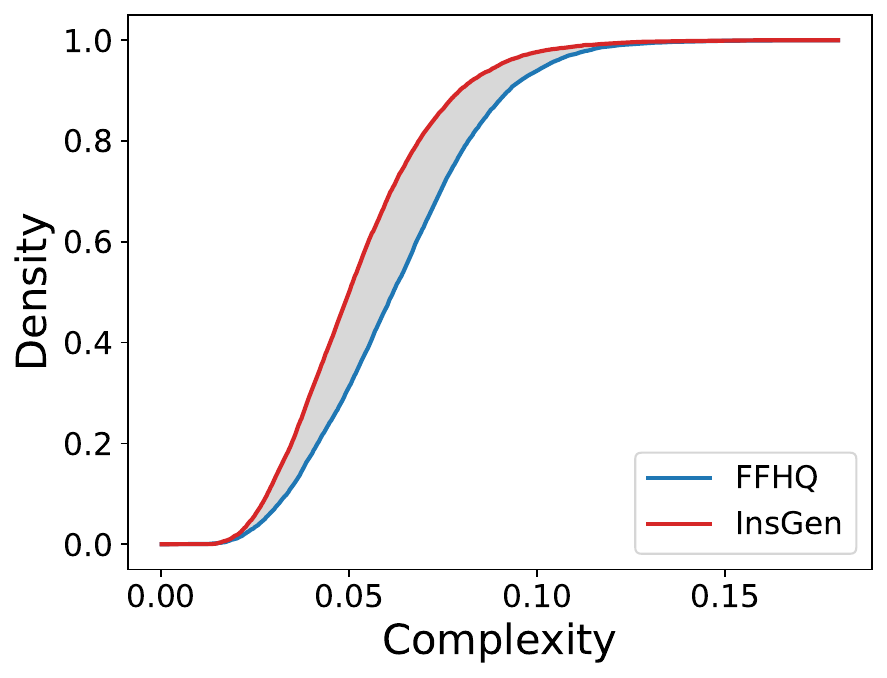}
\end{subfigure}
\begin{subfigure}[b]{0.32\columnwidth}
    \centering
    \includegraphics[width=\columnwidth]{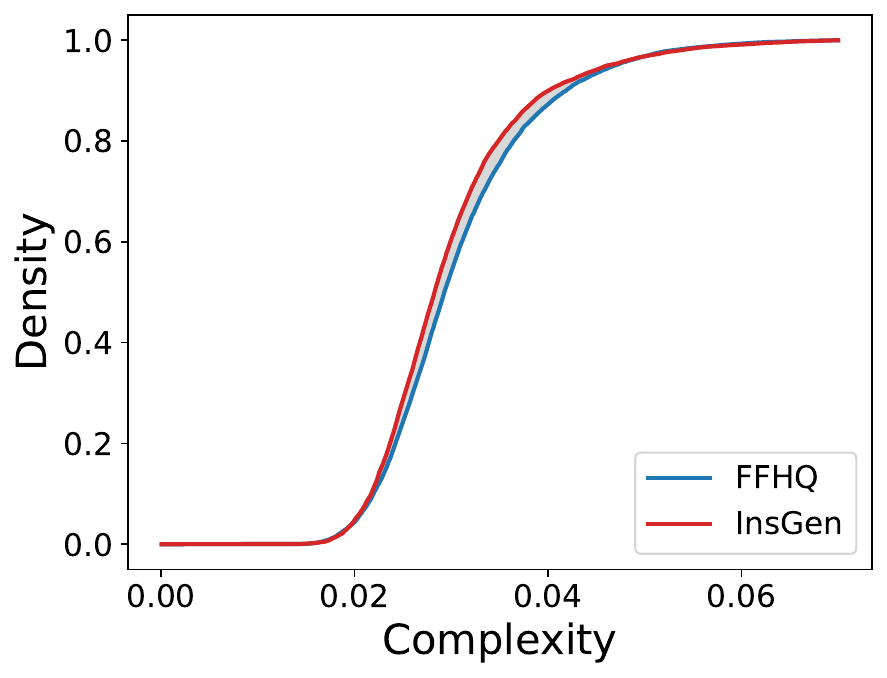}
\end{subfigure}
\begin{subfigure}[b]{0.32\columnwidth}
    \centering
    \includegraphics[width=\columnwidth]{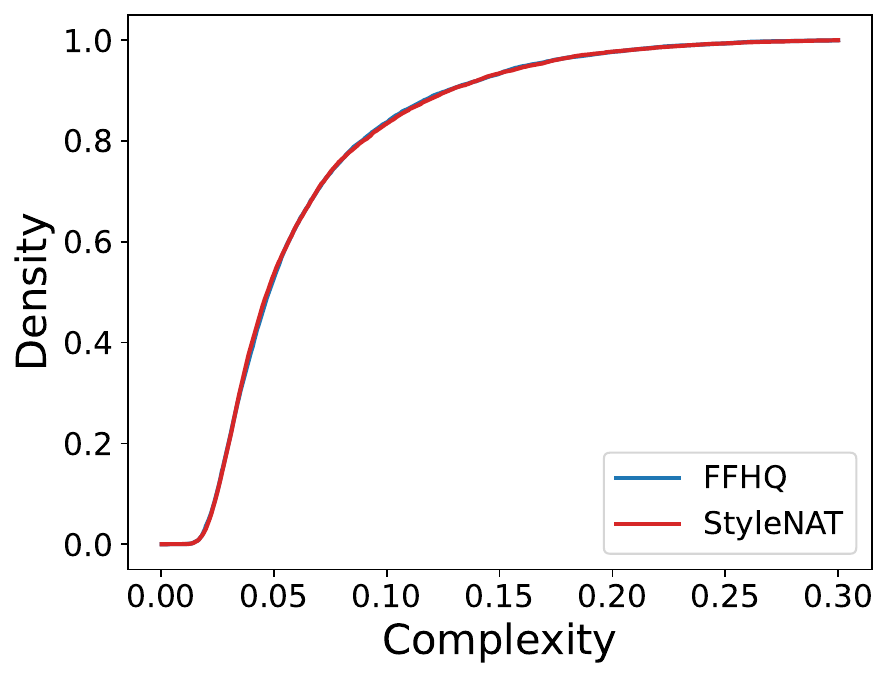}
\end{subfigure}
\begin{subfigure}[b]{0.32\columnwidth}
    \centering
    \includegraphics[width=\columnwidth]{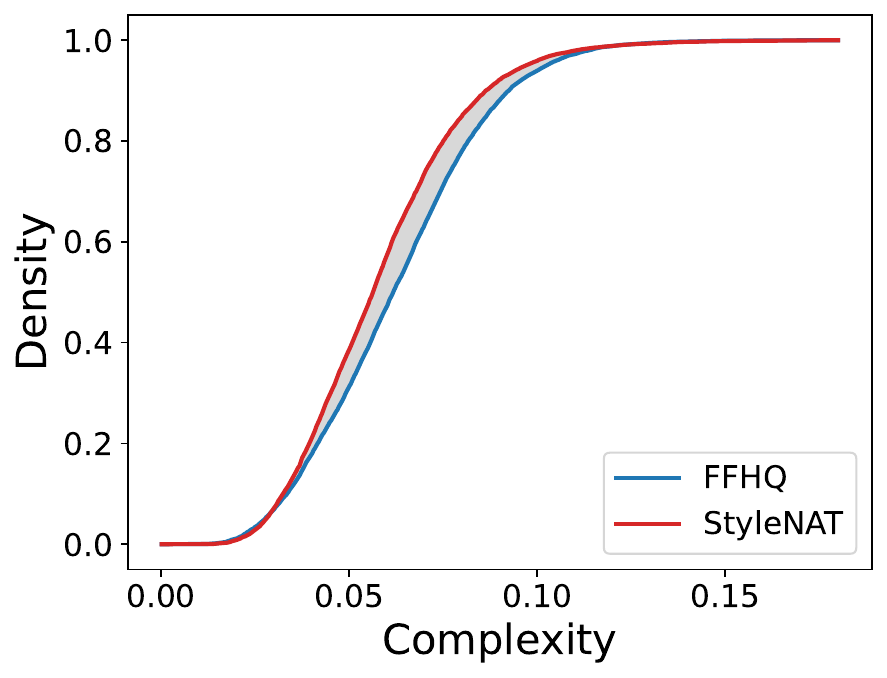}
\end{subfigure}
\begin{subfigure}[b]{0.32\columnwidth}
    \centering
    \includegraphics[width=\columnwidth]{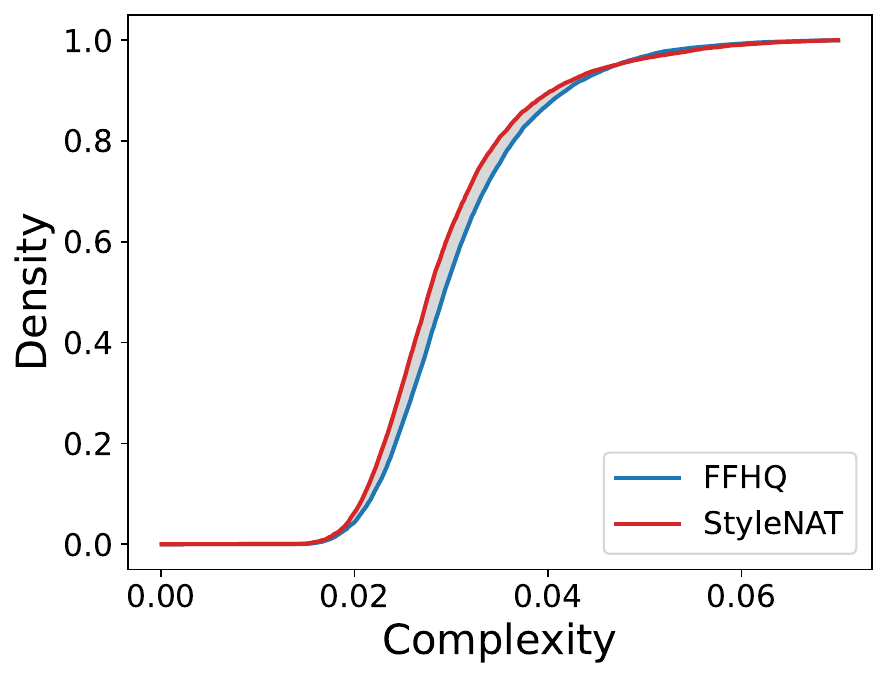}
\end{subfigure}
\caption{\textbf{Distribution of \com.} The cumulative distribution function (CDF) of \com is depicted for various feature models and generative models trained on FFHQ.
Each row shows distributions for each feature model: ViT-S (left), ConvNeXt-tiny (mid), and DINO-V2 (right).
Each column indicates a different type of generative model: InsGen \cite{insgen} (top) and StyleNAT \cite{stylenat} (bottom). Note that InsGen is assessed as a low-performance model compared to StyleNAT by the human evaluation \cite{dgm-eval}.
}
\label{fig:ffhq-complexity-ditribution}
\vspace{-1em}
\end{figure}

{\noindent \bf Results.}
\cref{tab: angle} shows the average values of the \com of the reference dataset and the generated dataset. Overall, the \com of the generated images is smaller than that of the reference images, which is confirmed by statistical tests, indicating that as expected, generated images are located in less complex regions in the representation space than the reference (real) images.
\cref{fig:ffhq-complexity-ditribution}  presents the cumulative distribution function (CDF) of \com for various feature models and datasets.
In most cases, the distribution of the generated dataset differs from that of the reference dataset.
Furthermore, the datasets generated by the InsGen model (left column), which generates relatively low-quality images, have more deviated distributions of \com compared to the datasets generated by the StyleNAT model, which generates higher-quality images.
The difference between the original dataset and the generated one is especially prominent in the case of using the ConvNeXt-tiny feature model.



\subsection{Vulnerability}
\label{sec:vulnerability}

\begin{figure}
\centering
\includegraphics[width=1.0\linewidth]{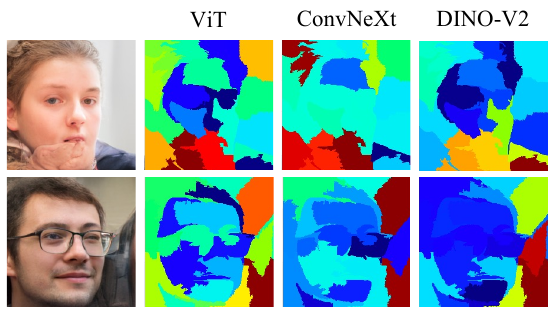}
\caption{\textbf{Image components causing large changes by adversarial attack.} 
We partition images into super-pixels and assess their contribution to feature changes by the attack. 
Starting from the left: the original image, and the level of contributions on the changes in the feature extracted by ViT-S \cite{vit}, ConvNeXt-tiny \cite{convnext}, and DINO-V2 \cite{dinov2}, respectively.
Red denotes a high level of impact on the changes, while blue indicates a low level of influence on the changes. 
}
\label{fig:explain}
\end{figure}

{\noindent \bf Motivation.}
We exploit the idea of adversarial attack \cite{fgsm, pgd, autoattack} as another tool for examining the relationship between the input space and the representation space.
In particular, we generate adversarial perturbations that cause large changes in the representation space with small changes in the input space.
As a result, we observe that unnatural components (regions) of images tend to cause large changes. 
The left column of \cref{fig:explain} shows examples of unnatural components of images, specifically the chin of the girl (upper panel) and the right side of the man (lower panel).
For each image, we employ the SLIC algorithm \cite{slic} to divide the image into 20 super-pixels.
Then, we randomly select 3 to 6 super-pixels among 20 super-pixels, add adversarial perturbations determined by PGD~\cite{pgd} to the selected super-pixels, and obtain the feature of the attacked image from a feature model.
We repeat this process 20 times. 
We apply linear regression between the binary variables indicating whether each super-pixel is attacked and the amount of feature change, and the obtained coefficient of each variable is considered as the contribution of each super-pixel to the feature change (see the appendix for more details).
The second to fourth columns of \cref{fig:explain} show the contribution of each super-pixel to the feature changes for different feature models with the color coding. 
Across all models, unnatural super-pixels are consistently highlighted in red, i.e., the feature of the image is largely changed when we add adversarial perturbations to the unnatural super-pixels.
Based on this result, we consider that examining the feature change of an image under adversarial attack can be an effective way to identify unnaturalness of the image.

\renewcommand{\arraystretch}{1}
\begin{table}[t]
\centering
\small
\begin{tabular}{cc|ccc}
\toprule
&&ViT&ConvNeXt&DINO-V2\\
\midrule
\multirow{3}{*}{CIFAR10}& Reference&24.56	&	14.57	&	29.41\\
&Generated&24.98	&	15.24	&	30.32\\
&$p$-value&$<$0.0001$^*$&$<$0.0001$^*$&$<$0.0001$^*$\\
\midrule
\multirow{3}{*}{ImageNet}& Reference&11.73	&	8.69	&	8.06\\
&Generated&15.80	&	12.45	&	12.85\\
&$p$-value&$<$0.0001$^*$&$<$0.0001$^*$&$<$0.0001$^*$\\
\midrule
\multirow{3}{*}{FFHQ}& Reference&18.30	&	14.57	&	12.90\\
&Generated&19.22	&	17.21	&	16.34\\
&$p$-value&$<$0.0001$^*$&$<$0.0001$^*$&$<$0.0001$^*$\\

\bottomrule
\end{tabular}
\caption{\textbf{\Vul of various datasets.} We compare the average value of \vul for various feature models, ViT-S \cite{vit}, ConvNeXt-tiny \cite{convnext}, and DINO-V2 \cite{dinov2}. `Reference' indicates the original dataset, such as CIFAR10 \cite{cifar}, ImageNet \cite{imagenet}, and FFHQ \cite{ffhq}. `Generated' denotes the \vul obtained from datasets generated by PFGM++ \cite{pfgm++}, RQ Transformer \cite{rqtrans}, and InsGen \cite{insgen} trained with the respective reference datasets.
`$p$-value' denotes the $p$-value of the one-tailed $t$-test under the null hypothesis that \vul of the generated dataset is equal to that of the reference dataset. The cases with statistical significance are marked with `$*$'.
}
\label{tab: vulnerability}
\vspace{-1em}
\centering
\end{table}

{\noindent \bf Definition.}
The PGD attack~\cite{pgd} is a widely used method for altering prediction results of a model through iterative perturbations applied to images.
While the original PGD attack targeting classification models aims to maximize the cross-entropy loss, we maximize the $L_2$ loss between the features of the original and attacked images.
Starting from $\mathrm{x}^0=\mathrm{x}+\delta \mathrm{N}$, the image is iteratively perturbed as follows.
Let $\mathrm{P}^{j}$ and $\mathrm{x}^{j}$ denote the adversarial perturbation and the attacked image at the $j$-th step of the attack, respectively. 
The modified PGD update rule is given by
\begin{equation}
    \mathrm{\Tilde{P}}^j = \nabla_{\mathrm{x}^{j}}{ L_2( M(\mathrm{x}), M(\mathrm{x}^j) ) },
\label{eq:our-attack}
\vspace{-1.5em}
\end{equation}
\begin{equation}
    \mathrm{P}^j = \alpha~ \mathrm{\Tilde{P}}^j / ||\mathrm{\Tilde{P}}^j||,
\label{eq:our-attack}
\vspace{-1.2em}
\end{equation}
\begin{equation}
	{\mathrm{x}}^{j+1} =
	\mathrm{Clip}_{0, 255} ( \mathrm{x}^{j} +  \mathrm{P}^j),
\end{equation}
where $\alpha$ is the attack size at each step, $M(\cdot)$ indicates the feature model, and $L_2( M(\mathrm{x}), M(\mathrm{x}^j))$ is the $L_2$ loss between the features of the original and modified inputs. $\mathrm{Clip}_{0, 255}(\cdot)$ is a clipping function that limits values to the range between 0 and 255.
Then, we define \vul of image $\mathrm{x}$, $\mathrm{V}(\mathrm{x})$, as follows:
\vspace{-.2em}
\begin{equation}
    \mathrm{V}(\mathrm{x}) = dist(M(\mathrm{x}), M(\mathrm{x}^J)),  
\label{eq:vul-j}
\vspace{-.2em}
\end{equation}
where $J$ is the total number of steps of the attack and $dist(A,B)$ indicates the $L_2$ distance between two vectors $A$ and $B$.

{\noindent \bf Experimental setup.}
We conduct an experiment to examine the validity of the \vul for evaluating unnaturalness of generated images.
We set $\alpha=0.01$, $\delta=10^{-6}$, and $J=10$ in this experiment, while keeping all other settings, such as feature models and generative models, the same as those used in \cref{sec:complexity}.

\begin{figure}
\centering
\begin{subfigure}[b]{0.32\columnwidth}
    \centering
    \includegraphics[width=\columnwidth]{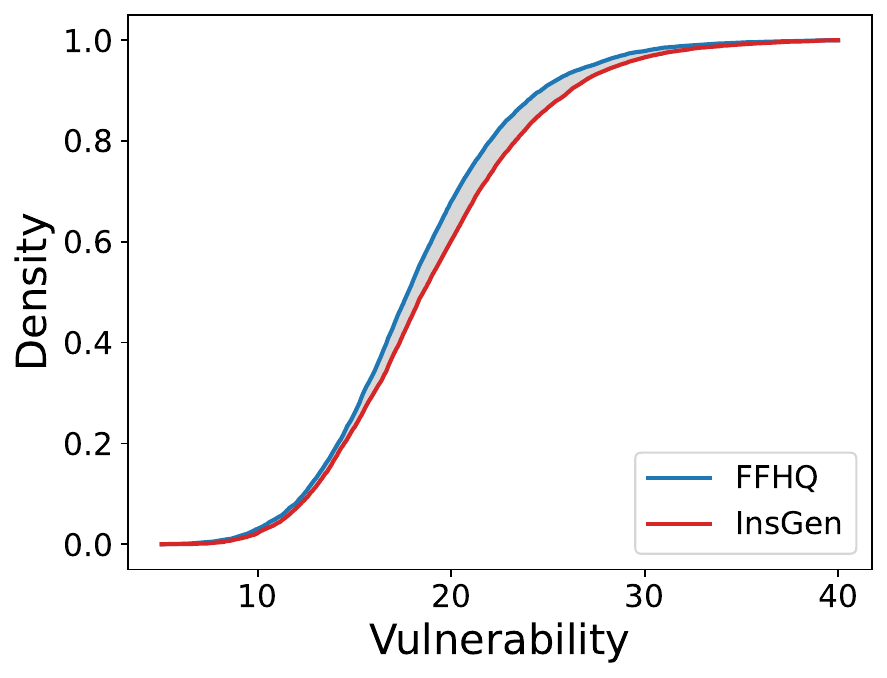}
\end{subfigure}
\begin{subfigure}[b]{0.32\columnwidth}
    \centering
    \includegraphics[width=\columnwidth]{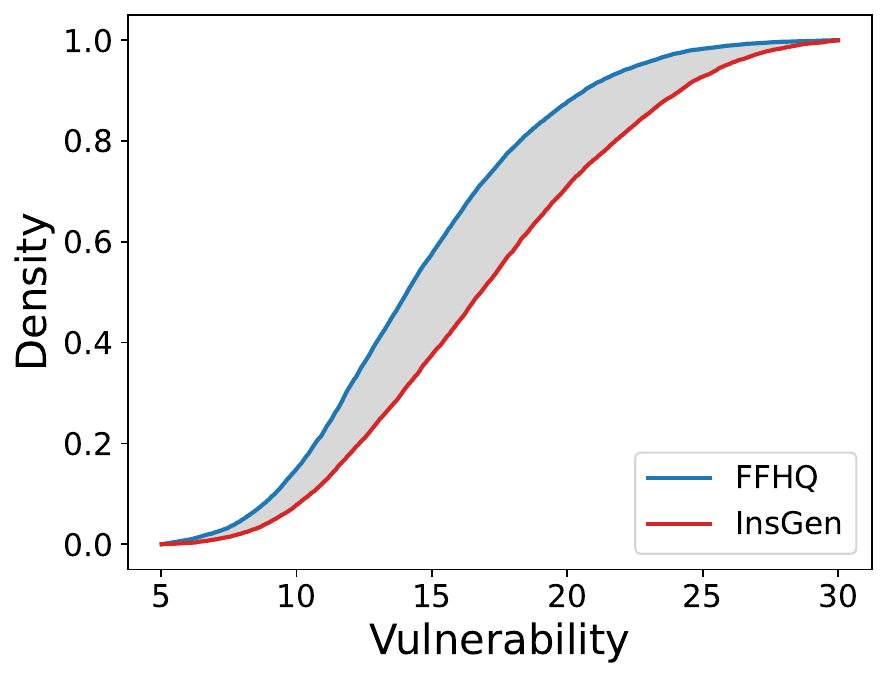}
\end{subfigure}
\begin{subfigure}[b]{0.32\columnwidth}
    \centering
    \includegraphics[width=\columnwidth]{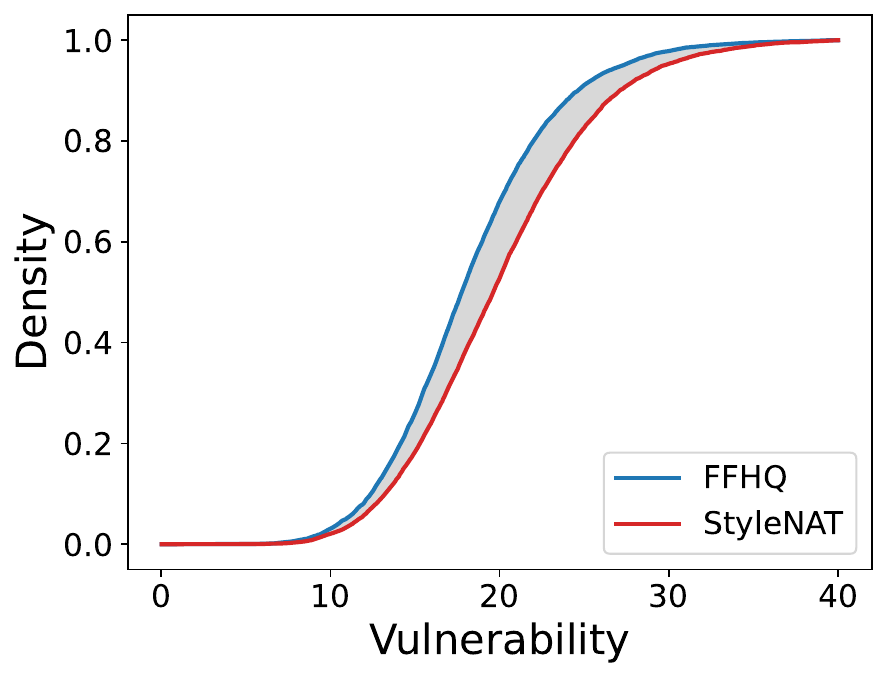}
\end{subfigure}
\begin{subfigure}[b]{0.32\columnwidth}
    \centering
    \includegraphics[width=\columnwidth]{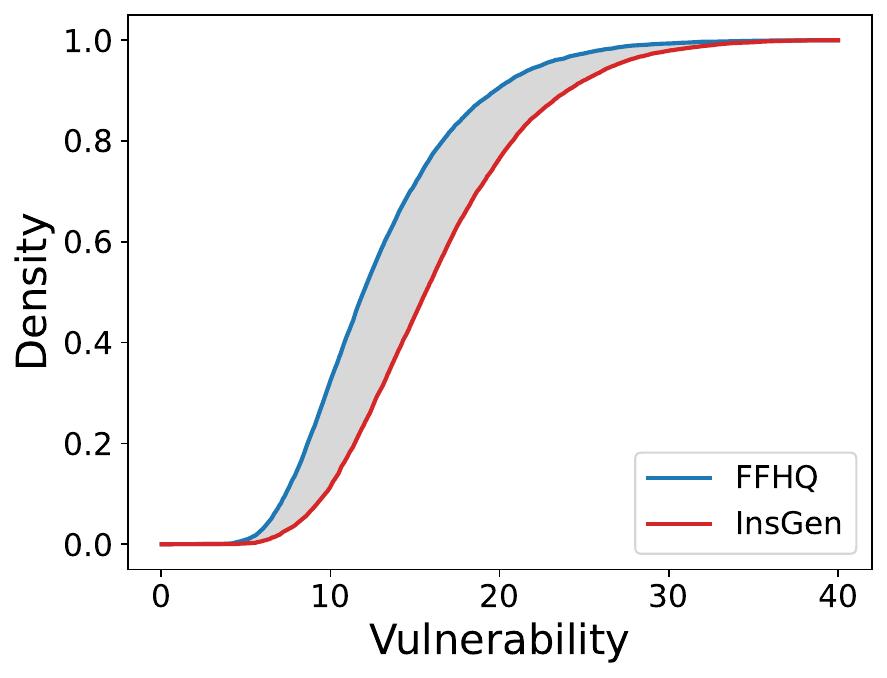}
\end{subfigure}
\begin{subfigure}[b]{0.32\columnwidth}
    \centering
    \includegraphics[width=\columnwidth]{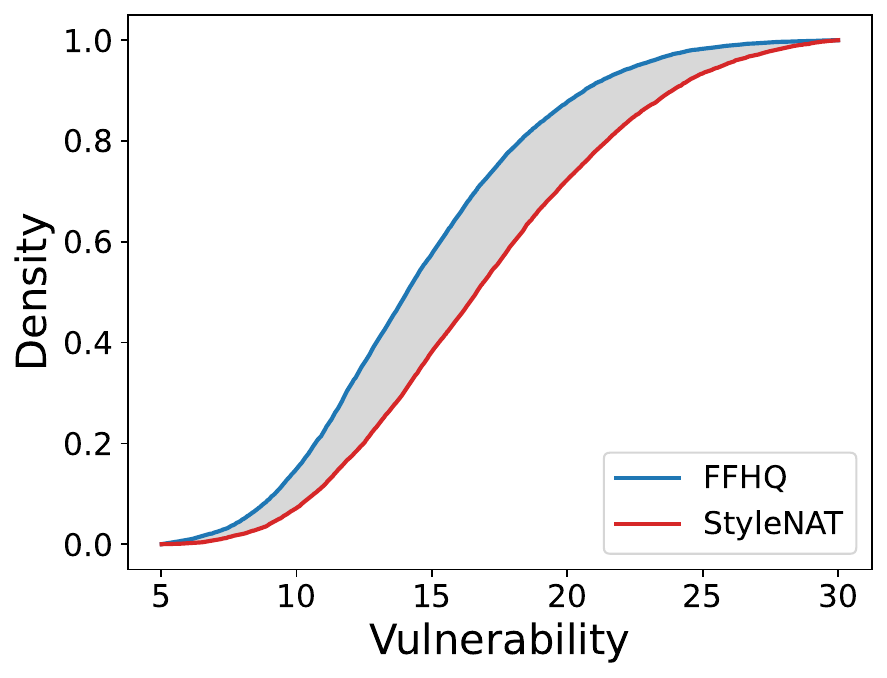}
\end{subfigure}
\begin{subfigure}[b]{0.32\columnwidth}
    \centering
    \includegraphics[width=\columnwidth]{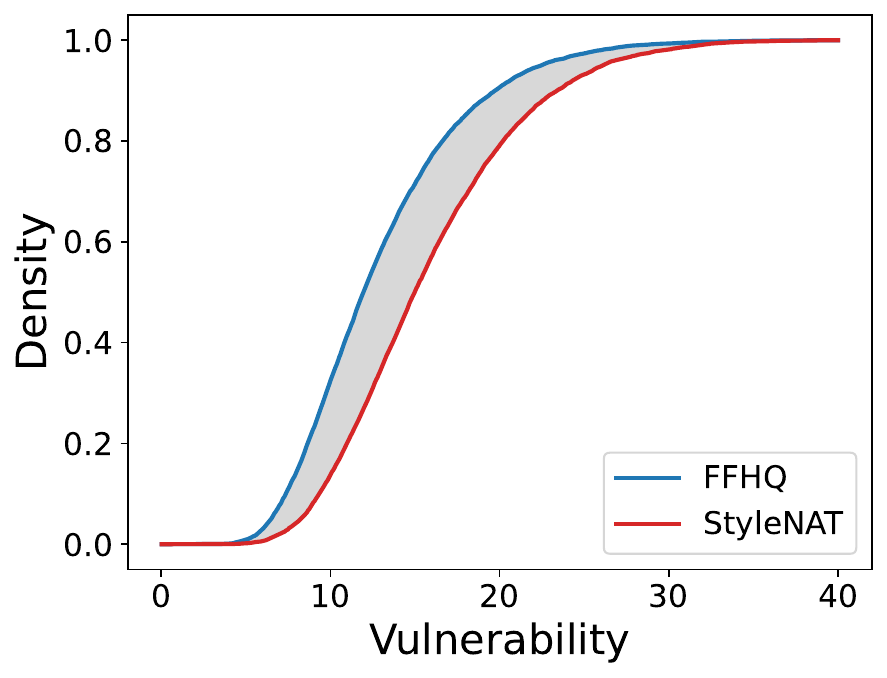}
\end{subfigure}
\caption{\textbf{Distribution of \vul.} The cumulative distribution function (CDF) of \vul is depicted for various feature models and generative models trained on FFHQ.
Each row shows distributions for each feature model: ViT-S (left), ConvNeXt-tiny (mid), and DINO-V2 (right).
Each column indicates a different type of generative model: InsGen \cite{insgen} (top) and StyleNAT \cite{stylenat} (bottom). Note that InsGen is assessed as a low-performance model compared to StyleNAT by the human evaluation \cite{dgm-eval}.}
\label{fig:ffhq-vulnerability-ditribution}
\end{figure}

{\noindent \bf Results.} 
\cref{tab: vulnerability} shows the average \vul of images from various datasets on ViT-S, ConvNeXt-tiny, and DINO-V2 as feature models.
Additional results for other models are shown in the appendix.
It can be seen that the \vul of generated images is larger than that of reference images.
\cref{fig:ffhq-vulnerability-ditribution} also shows the significant disparity in the \vul distribution between real and generated images.
Furthermore, the high-quality generated dataset by the StyleNAT model has a more similar distribution to that of the reference dataset compared to the relatively low-quality generated dataset by the InsGen model in most cases, which is consistent with the results observed in the comparison of \com.



\begin{figure}
\centering
\begin{subfigure}[b]{0.48\columnwidth}
    \includegraphics[width=\columnwidth]{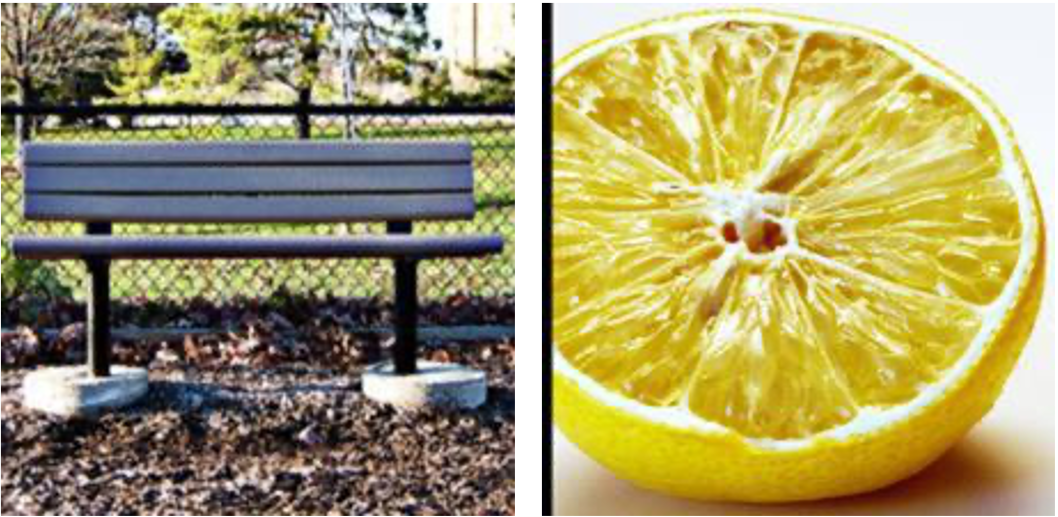}
    \caption{Low-MVT level}
    \label{fig:flash-ex-a}
\end{subfigure}
\begin{subfigure}[b]{0.48\columnwidth}
    \includegraphics[width=\columnwidth]{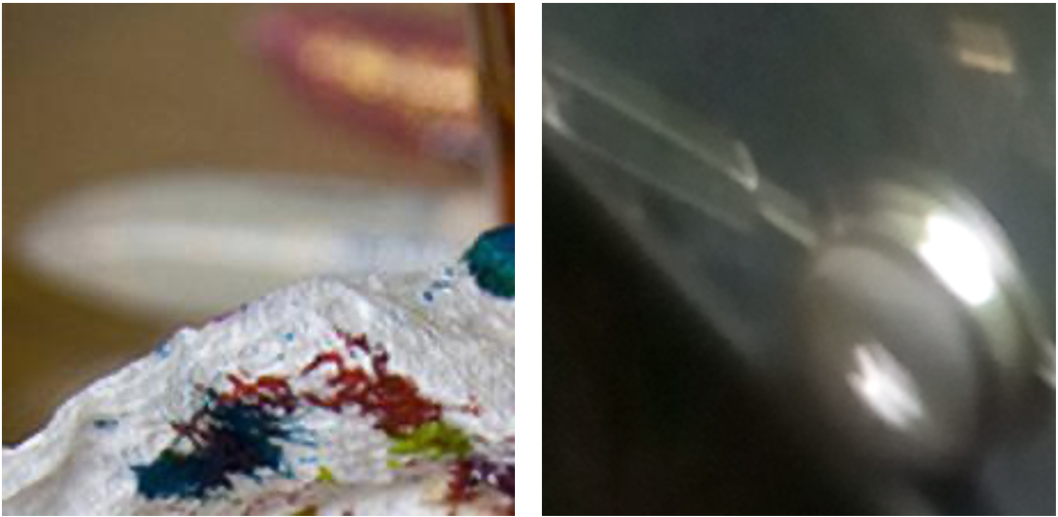}
    \caption{High-MVT level}
    \label{fig:flash-ex-b}
\end{subfigure}
\caption{\textbf{Examples of the MVT dataset.} Images having low MVT levels are clear for people to recognize. On the other hand, a high MVT level means that people need a long time to recognize the image contents due to weak naturalness.}
\label{fig:flash-ex}
\vspace{-1em}
\end{figure}

\begin{figure}
\centering
\includegraphics[width=0.5\linewidth]{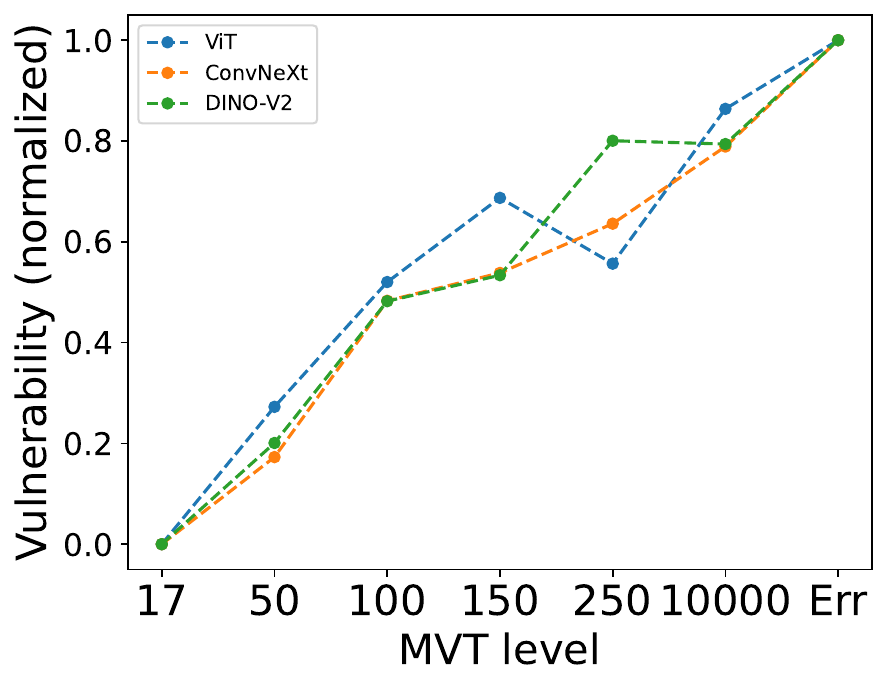}
\caption{\textbf{\Vul vs. MVT levels.} The average value of \vul of images having each MVT level. Note that the value of \vul is normalized by the maximum and the minimum values in each feature model.}
\label{fig:flash}
\end{figure}

{\noindent \bf \Vul and naturalness.}
We further demonstrate the relationship between \vul and naturalness. 
We employ the MVT dataset \cite{mvt}, which contains information on the difficulty for humans to recognize objects in each image by measuring the minimum viewing time required for recognition.
In the dataset, the viewing time is separated into several levels called MVT levels (i.e., 17 ms, 50 ms, 100 ms, 150 ms, 250 ms, and 10,000 ms).
Each MVT level means the time within which over 50\% of participants can correctly classify an object. 
It is linked to the level of unnaturalness of the images. In other words, when an image contains clear content (\cref{fig:flash-ex-a}), humans swiftly identify the depicted object within 50 ms. Conversely, an image with unnatural components (\cref{fig:flash-ex-b}) poses a challenge for human recognition, requiring over 10,000 ms.

\cref{fig:flash} shows the \vul with respect to the MVT level utilizing the MVT dataset.
Here, ``Err'' indicates that over 50\% of participants misclassify an image given sufficient time.
The MVT level is highly correlated with the \vul on all feature models, i.e., low naturalness is related to high \vul. 


%% file: sec/4_metric.tex
\section{Evaluating generative models}
\label{sec:metric}

In the previous section, we explored the distinct properties in the representation space around generated images in comparison to those of real images.
The representation space around generated images is more complex (\com) and contains a certain path that is vulnerable to adversarial changes (\vul).
Thus, we propose a novel metric for evaluating generative models by capturing anomalies based on both \com and \vul, called anomaly score.

\subsection{Anomaly score for generative models}
We define anomaly score (AS) for evaluating generative models as the difference of the bivariate distributions of \com and \vul between the reference and generated datasets.
We denote the set of anomaly vectors for the dataset $\mathrm{X}$ as $\textrm{A}(\mathrm{X})=\{[\textrm{C}(\mathrm{x}),\textrm{V}(\mathrm{x})]\}_{\mathrm{x} \in \mathrm{X}}$.

To compare the distributions of the generated dataset and the reference dataset, we employ the Kolmogorov-Smirnov (KS) statistic that measures a non-parametric statistical difference between two distributions. Our anomaly score, AS, utilizing the 2D KS statistic, is defined as
\begin{equation}
    \textrm{AS} = \textrm{Sup}~\big |\textrm{CDF}(\textrm{A}(\mathrm{X}^{real}))-\textrm{CDF}(\textrm{A}(\mathrm{X}^{generated}))\big |,
\end{equation} 
where $\mathrm{X}^{real}$ is the reference dataset for the generated dataset, $\mathrm{X}^{generated}$, and $\textrm{CDF}(\cdot)$ is the cumulative distribution function of input vectors. 
We compute 2D KS statistics by referencing \url{github.com/syrte/ndtest}.
AS has the minimum value of 0 when the two distributions are identical and the maximum value of 1 when the two distributions are significantly different, i.e., their CDFs are completely non-overlapped.

\begin{figure}[t]
    \centering
    \begin{subfigure}[b]{0.45\linewidth}
        \includegraphics[width=\textwidth]{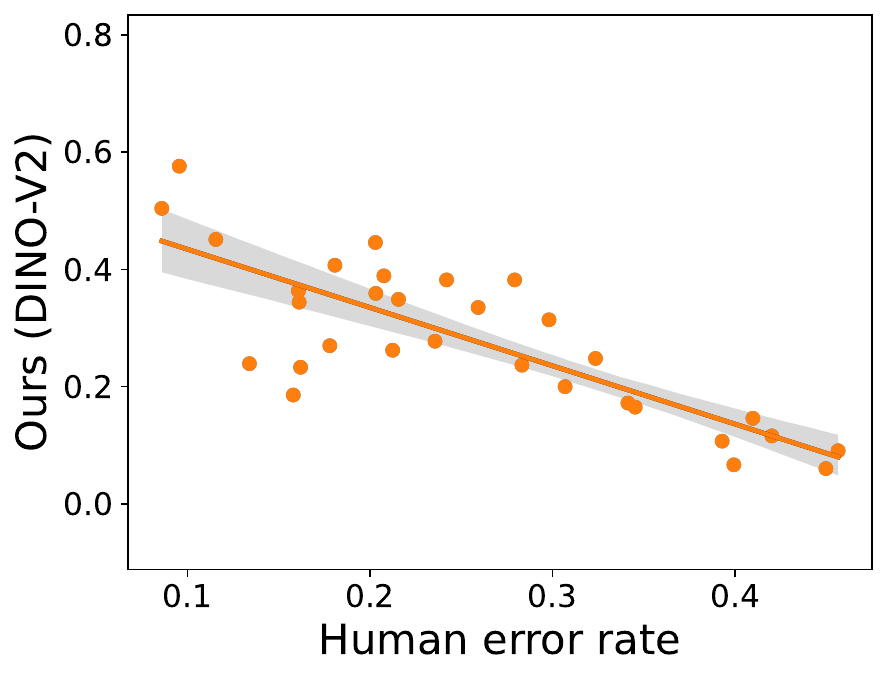}
        \vspace{-1em}
        \caption{Ours (DINO-V2)}
        \label{fig:perform-a}
    \end{subfigure}
    \begin{subfigure}[b]{0.45\linewidth}
        \includegraphics[width=\textwidth]{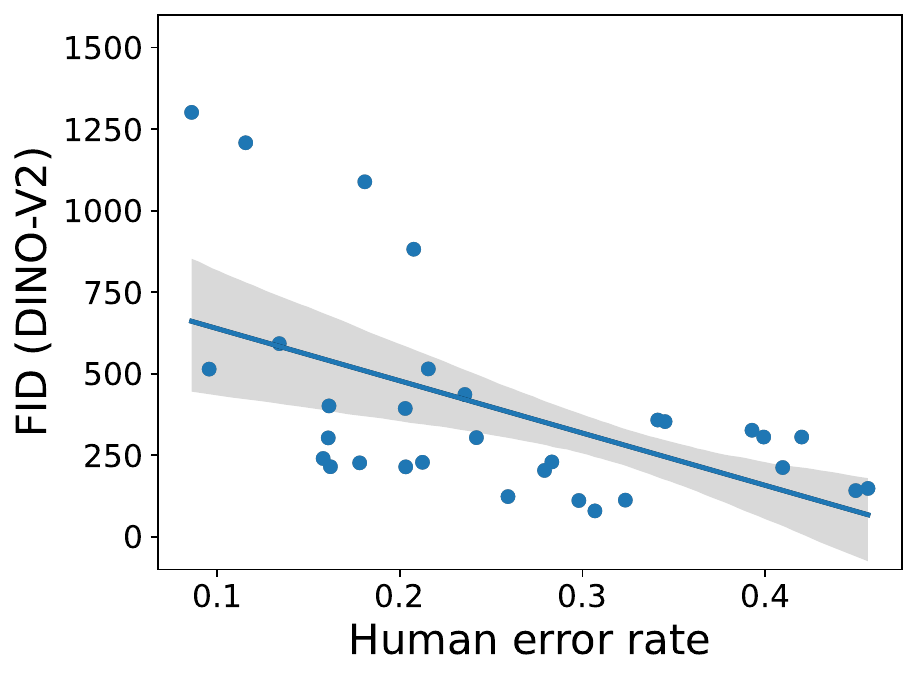}
        \vspace{-1em}
        \caption{FID (DINO-V2)}
        \label{fig:perform-b}
    \end{subfigure}
    \vspace{-.5em}
    \caption{\textbf{Performances of our method and FID using DINO-V2 for overall datasets.} Each dot represents a distinct dataset generated by a generative model. A high human error rate indicates a high-quality dataset, while a high AS score means a low-quality dataset.}
    \vspace{-.8em}
    \label{fig:perform}
\end{figure}

\begin{figure*}[t]
\centering
\begin{subfigure}[b]{\textwidth}
    \centering
    \includegraphics[width=0.24\columnwidth]{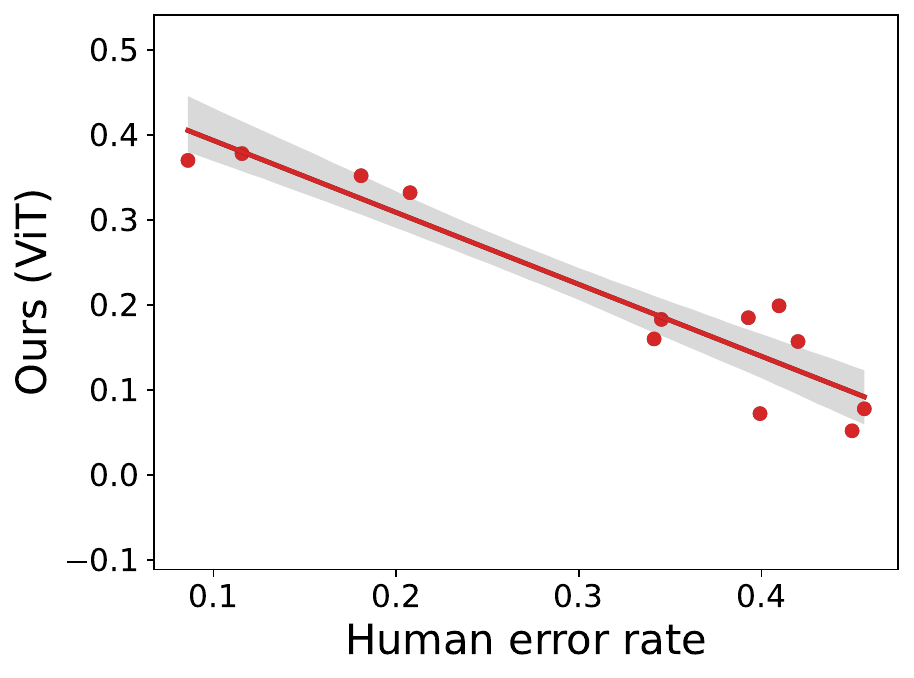}
    \includegraphics[width=0.24\columnwidth]{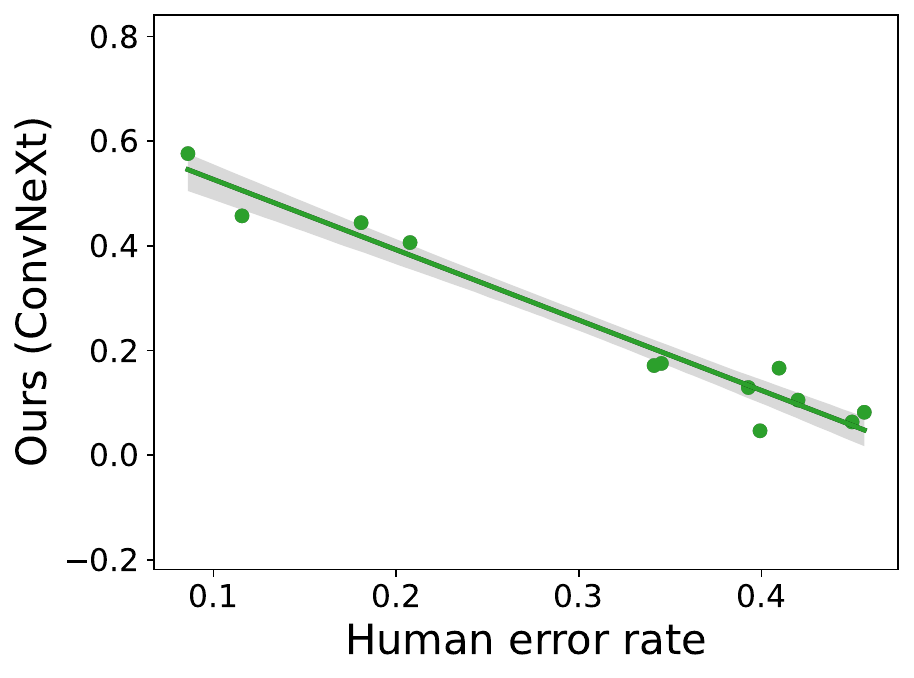}
    \includegraphics[width=0.24\columnwidth]{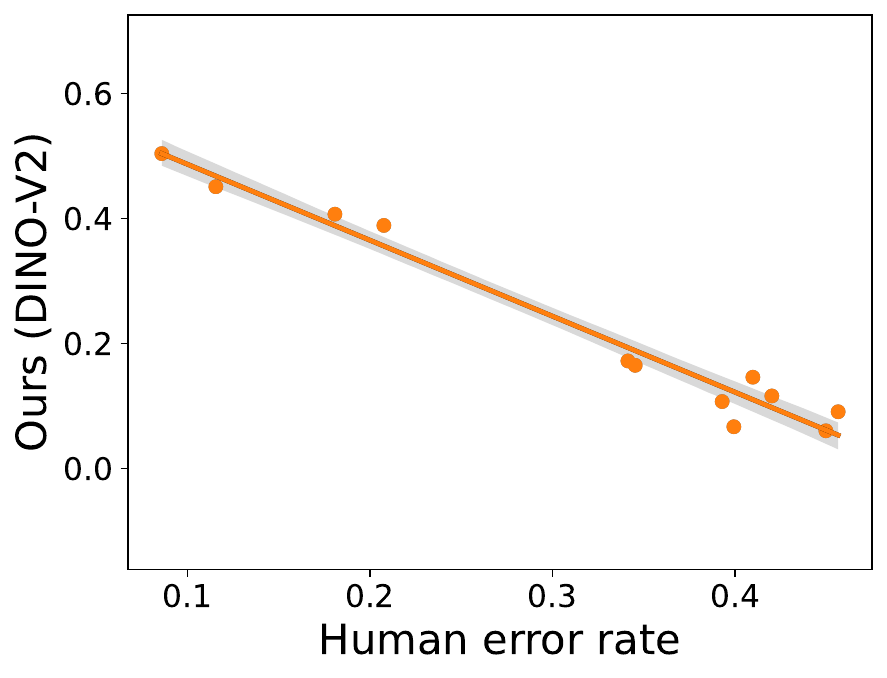}
    \includegraphics[width=0.24\columnwidth]{./figs/cifar_aa_fid.pdf}
    \caption{CIFAR10}
\end{subfigure}

\begin{subfigure}[b]{\textwidth}
\centering
    \includegraphics[width=0.24\columnwidth]{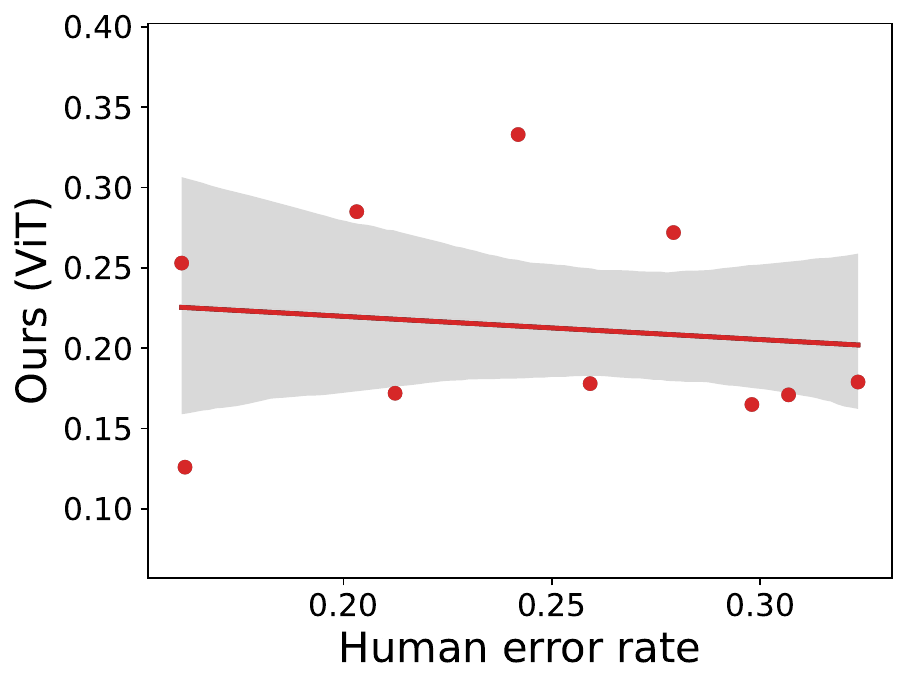}
    \includegraphics[width=0.24\columnwidth]{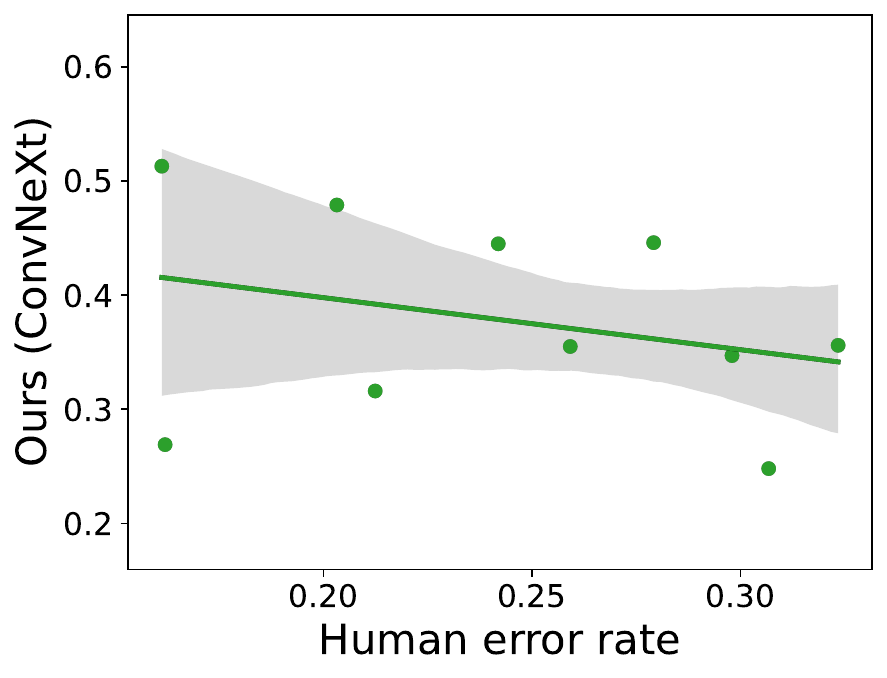}
\includegraphics[width=0.24\columnwidth]{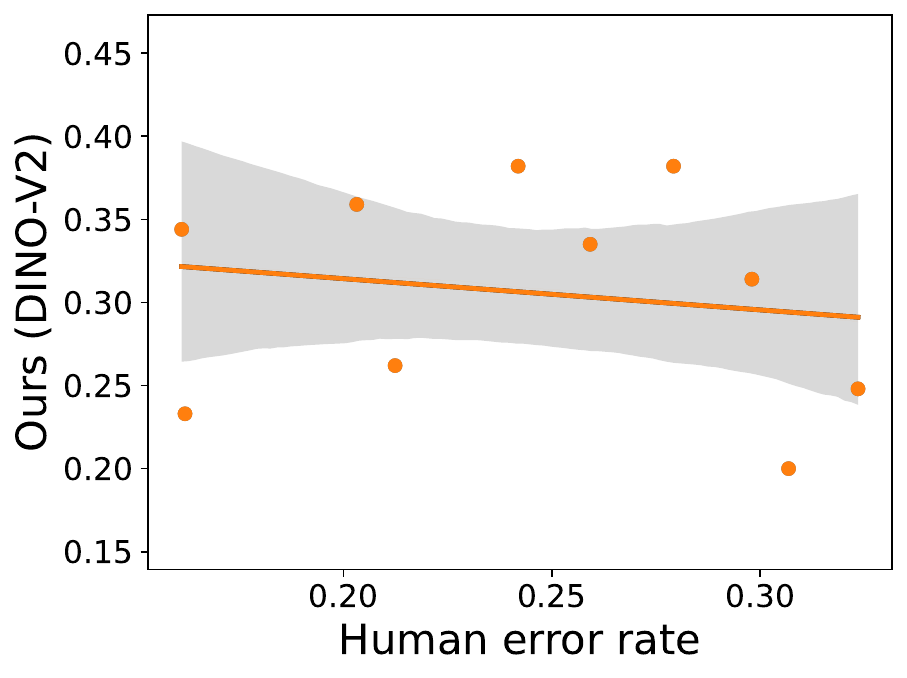}
\includegraphics[width=0.24\columnwidth]{./figs/imagenet_aa_fid.pdf}
    \caption{ImageNet}
\end{subfigure}

\begin{subfigure}[b]{\textwidth}
    \centering
    \includegraphics[width=0.24\columnwidth]{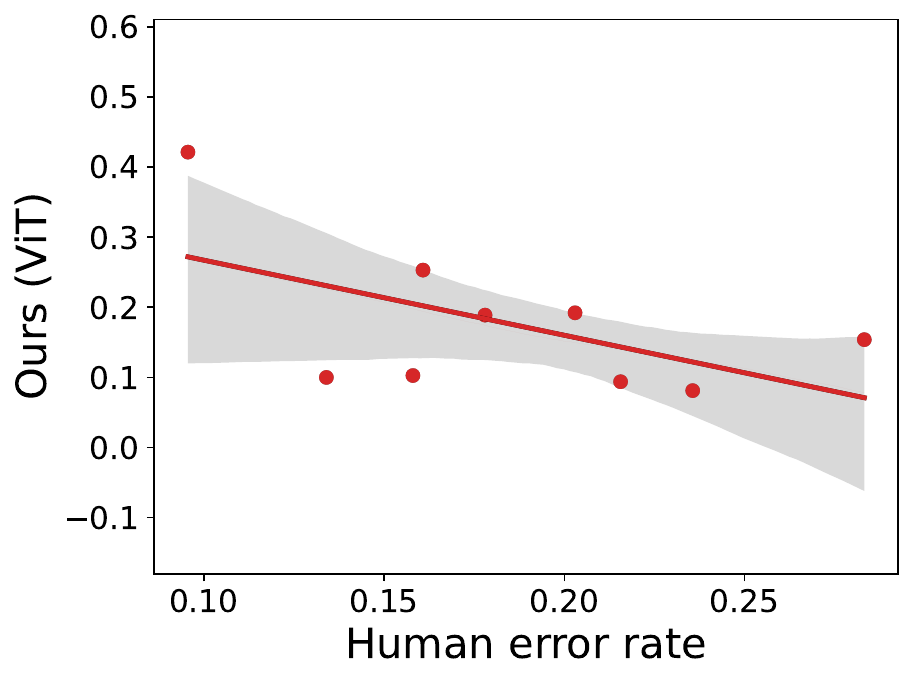}
\includegraphics[width=0.24\columnwidth]{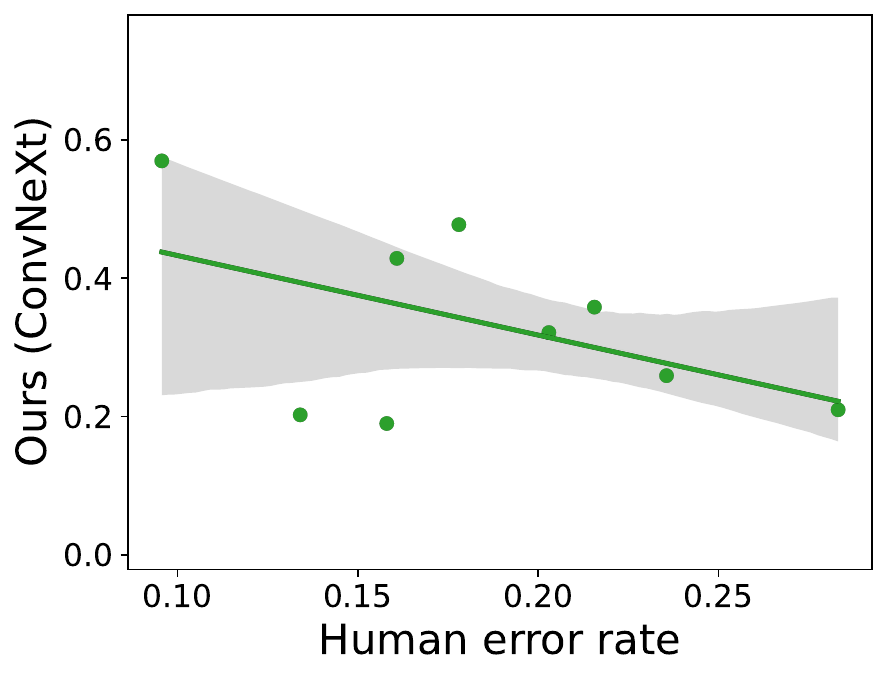}
\includegraphics[width=0.24\columnwidth]{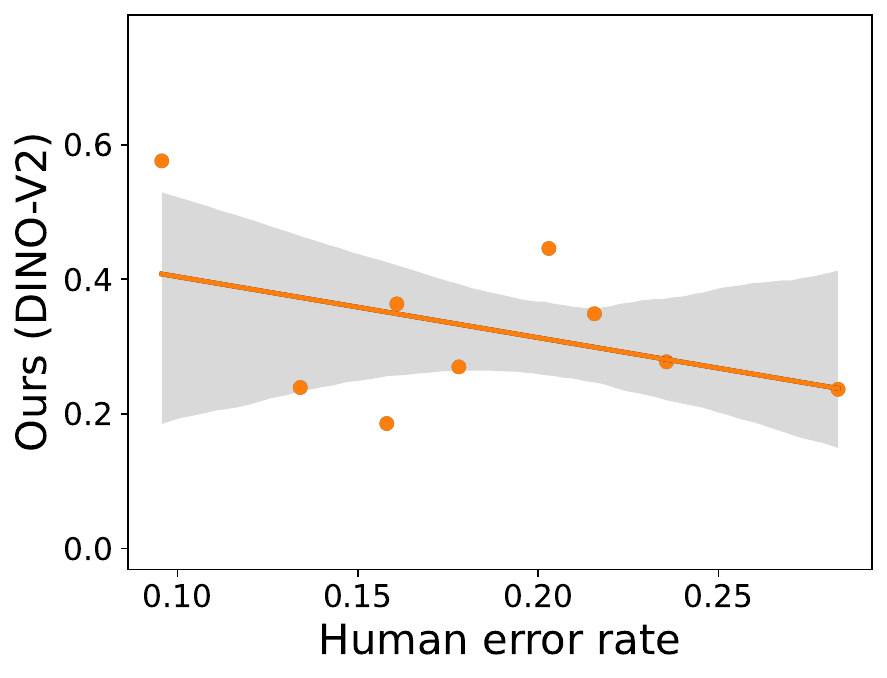}
\includegraphics[width=0.24\columnwidth]{./figs/ffhq_aa_fid.pdf}
    \caption{FFHQ}
\end{subfigure}
\vspace{-2em}
\caption{\textbf{Overall results of evaluating generative models on each dataset type.} Each dot represents a distinct dataset generated by a generative model. A high human error rate indicates a high-quality dataset, while a high AS score means a low-quality dataset. The first three columns show AS with different feature models: DINO-V2, ConvNeXt-tiny, and ViT-S, respectively. The last column is the result of FID~\cite{FID} with the DINO-V2 model.}
\label{fig:all-anomaly}
\vspace{-1em}
\end{figure*}

\begin{figure}
\centering
\begin{subfigure}[b]{0.32\columnwidth}
    \centering
    \includegraphics[width=\columnwidth]{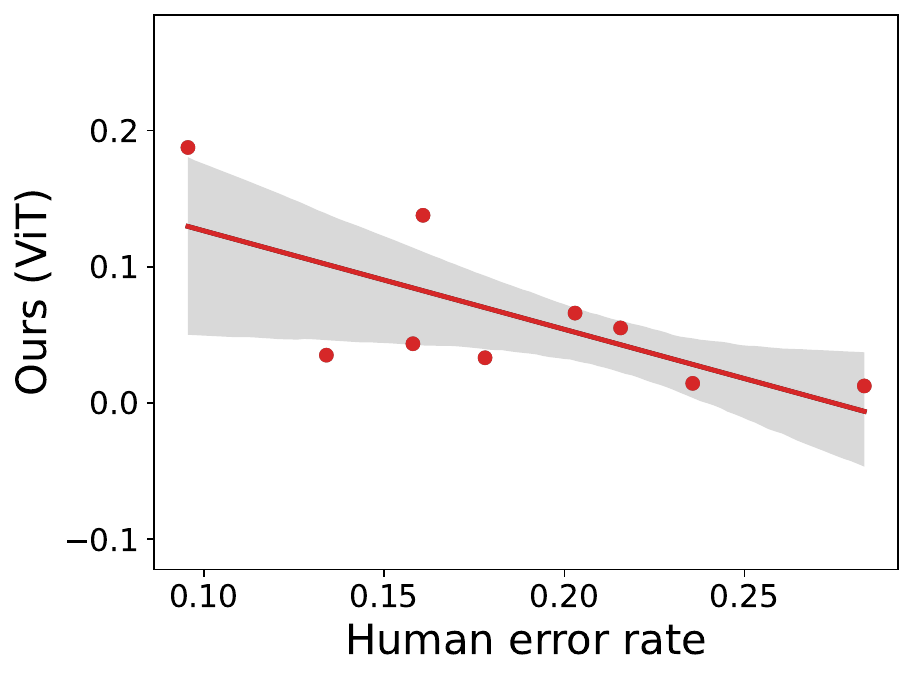}
\end{subfigure}
\begin{subfigure}[b]{0.32\columnwidth}
    \centering
    \includegraphics[width=\columnwidth]{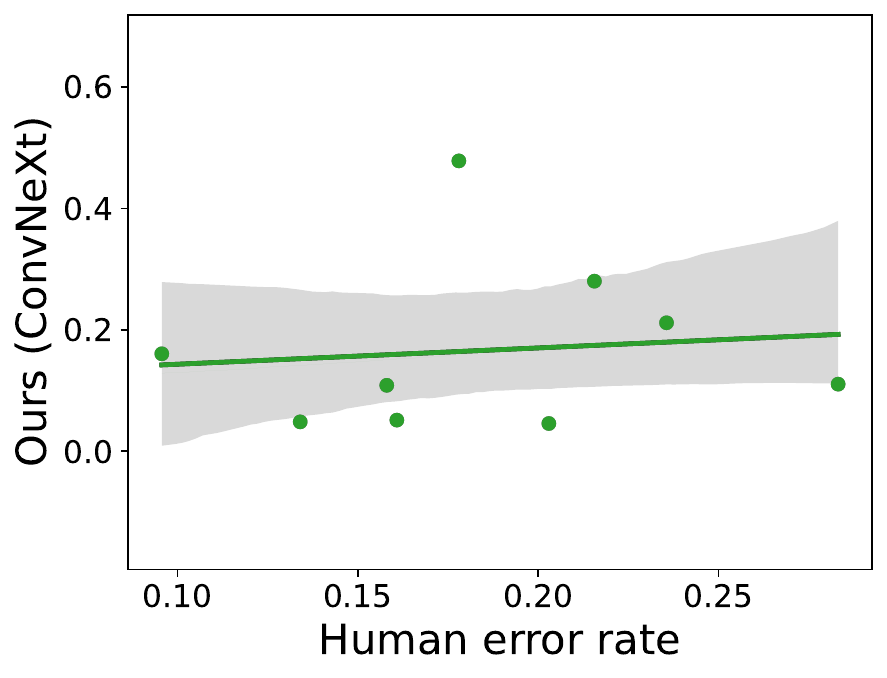}
\end{subfigure}
\begin{subfigure}[b]{0.32\columnwidth}
    \centering
    \includegraphics[width=\columnwidth]{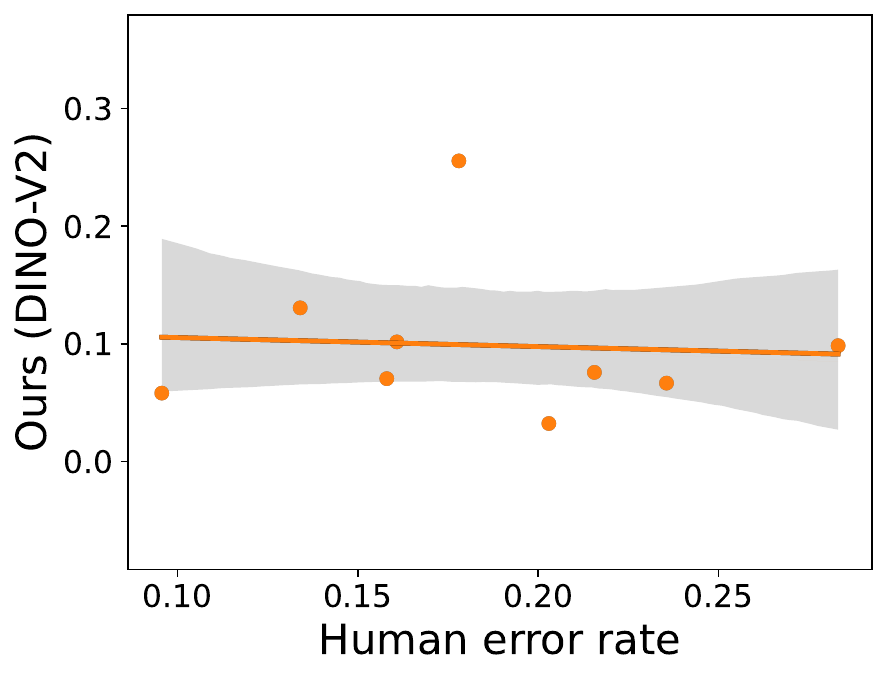}
\end{subfigure}
\begin{subfigure}[b]{0.32\columnwidth}
    \centering
    \includegraphics[width=\columnwidth]{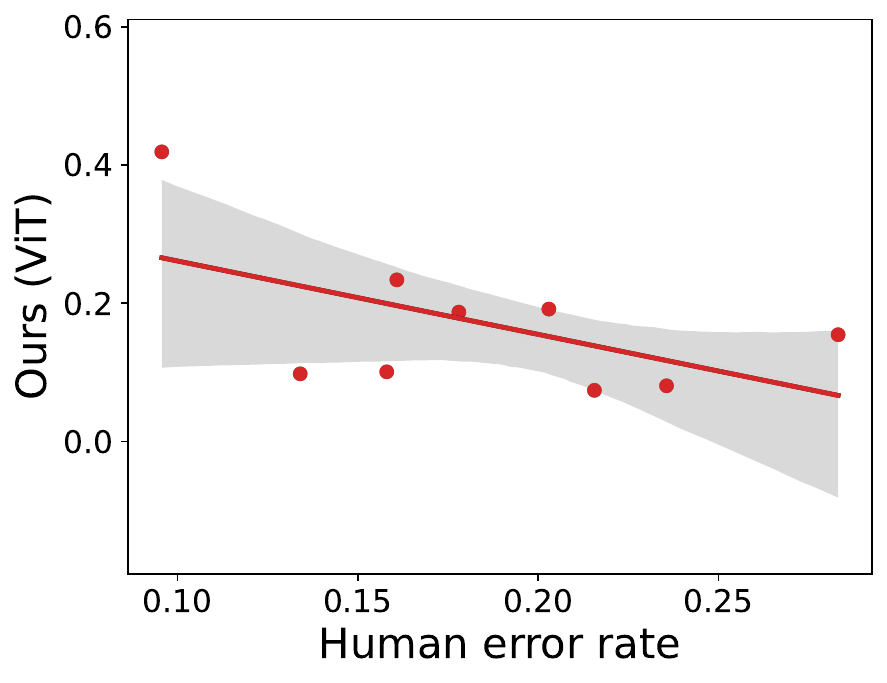}
\end{subfigure}
\begin{subfigure}[b]{0.32\columnwidth}
    \centering
    \includegraphics[width=\columnwidth]{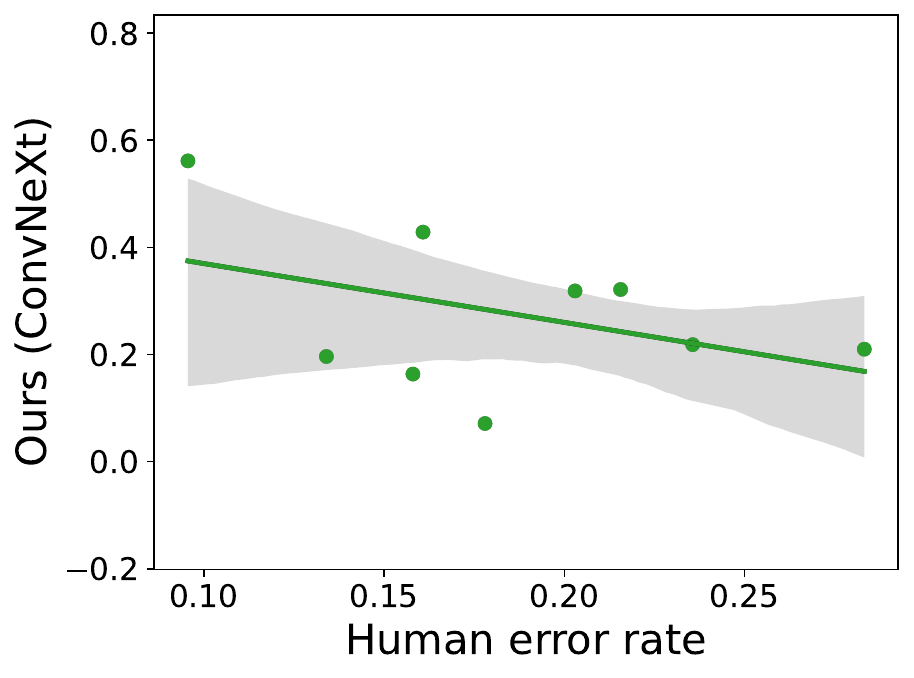}
\end{subfigure}
\begin{subfigure}[b]{0.32\columnwidth}
    \centering
    \includegraphics[width=\columnwidth]{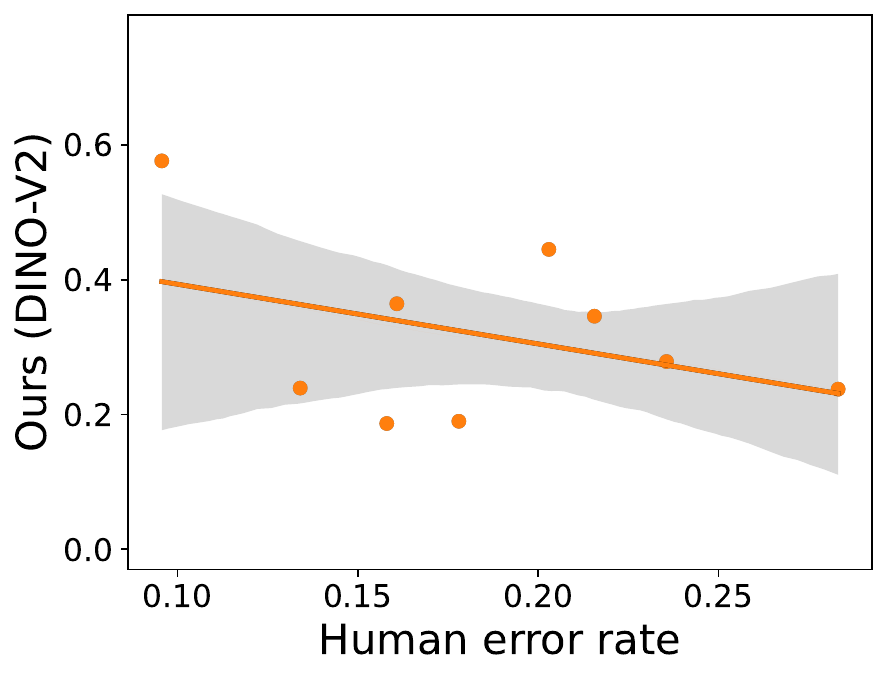}
\end{subfigure}
\caption{\textbf{Anomaly score using one measure.} Each dot represents a distinct generated dataset produced by a generative model for the FFHQ dataset. A high human error rate means a high-quality dataset, while a high AS implies a low-quality dataset.
Each column corresponds to a different feature model, and each row represents the results obtained by using the distribution of \com (top) or \vul (bottom). }
\label{fig:1d}
\vspace{-2em}
\end{figure}

\subsection{Experiments}
\label{sec:results}
{\noindent \textbf{Setup.}}
To evaluate the effectiveness of our metric, we examine whether our scores align with the subjective scores reported in \cite{dgm-eval}.
In this subjective test, human viewers responded whether a given image appeared fake (generated) or real.
Note that a high human error rate for a generated dataset indicates that many viewers cannot identify the images as fake ones, implying that the images in the dataset have high quality and are realistic.
We employ generated datasets produced by various generative models including GANs \cite{acgan, biggan, logan, mhgan, reacgan, sg, sg2, sgxl, wgan, styleswin, progan, gigagan, insgen, stylenat, maskgit}, VAEs \cite{vae, rqtrans}, a flow-based generative model \cite{resflow}, and diffusion models \cite{ddpm, iddpm, ldm, unleash, adm, ditxl2, admg, pfgm++, lsgm} from dgm-eval \cite{dgm-eval} and utilize 10000 generated images from each dataset. 
We set $K$ of \cref{eq:angle} and $J$ of \cref{eq:vul-j} to 10. During feature extraction, our method needs $K$-times inferences for computing \com and $J$-times inferences for computing \vul, while feature extraction FID needs one inference. In our setting, we need 20-times inferences.

{\noindent \textbf{Results.}} 
\cref{fig:perform} shows the comprehensive performance of our method with DINO-V2 as a feature model. It contains evaluation results on all generative models targeting one of the CIFAR10, ImageNet, and FFHQ datasets.
For comparison, the conventional FID with DINO-V2 as a feature model is also evaluated. 
We can observe that our method shows better performance than FID in terms of evaluating naturalness of generated images.
Our method has a relatively high correlation (-0.81 pearson correlation coefficient (PCC)) with human perception, while FID has a lower correlation (-0.54 PCC).

\cref{fig:all-anomaly} shows AS and FID with various feature models for evaluating generative models trained on different datasets.
Our method outperforms FID (the last column) on the FFHQ dataset. AS has a relatively high correlation (-0.56 PCC) with human perception, while FID has a lower correlation (-0.38 PCC).
We can observe that AS with DINO-V2 (-0.98 PCC), ConvNeXt-tiny (-0.30 PCC), and ViT-S (-0.56 PCC) are well aligned with the human error rate on CIFAR10, ImageNet, and FFHQ, respectively.
Additional results regarding the evaluation of generative models using other feature models are presented in the appendix.

\noindent \textbf{1D test.}
In our method, we utilize both \com and \vul of the images.
We conduct a comparative analysis using only one of the two in the anomaly vector, where the 1D KS statistic is used instead of the 2D KS statistic.
\cref{fig:1d} presents the results on the FFHQ dataset.
\Com performs well with ViT-S (-0.69 PCC) but not with ConvNeXt-tiny and DINO-V2 (0.11 and -0.07 PCCs, respectively). \Vul shows decent performance over all feature models (-0.55, -0.40, and -0.39 PCCs, respectively), but is outperformed by our method employing both \com and \vul (-0.56, -0.48, and -0.42 PCCs, respectively).
The results on the overall dataset are shown in \cref{tab: perform}.
We can observe that the 2D-metric shows better performance than the 1D-metrics as well as FID when we use DINO-V2 as a feature model.
These results demonstrate the benefit of employing both \com and \vul in our AS in terms of both performance and consistency.

\begin{table}[t]
\centering
\small
\begin{tabular}{c|cc}
\toprule
 & PCC & SRCC \\
\midrule
1D-\Com & -0.376 & -0.388 \\
1D-\Vul & -0.790 & -0.722 \\
2D (\textbf{Ours}) & \underline{-0.806} & \underline{-0.738} \\
FID & -0.540 & -0.525 \\
\bottomrule
\end{tabular}
\vspace{-.5em}
\caption{\textbf{Evaluation results of generated datasets.} We compare 
Pearson correlation coefficient (PCC) and Spearman rank correlation coefficient (SRCC) between the human error rate and the result of each metric by using DINO-V2 as a feature model.}
\vspace{-1.5em}
\label{tab: perform}
\end{table}

%% file: sec/5_individual_metric.tex
\section{Evaluating individual generated images}
\label{sec:metric-individual}

In this section, we propose a method for evaluating generated images individually based on \com and \vul defined in \cref{sec:why}.

\subsection{Anomaly score for individual generated images}
We adopt a simple and effective formula to capture the properties of \com and \vul of an image in a single score.
The proposed anomaly score for individual images (AS-i) is defined as follows:

\begin{equation}
    \textrm{AS-i}(\mathrm{x}) = \frac{\textrm{V}(\mathrm{x})}{\textrm{C}(\mathrm{x})},
\end{equation}
where $\mathrm{x}$ represents an individual generated image.
When the image is natural, \com around the image increases (large $\textrm{C}(\mathrm{x})$) and \vul of the image decreases (small $\textrm{V}(\mathrm{x})$), hence AS-i becomes small.
On the other hand, when the image is unnatural, AS-i becomes large.

\subsection{Subjective test} 

\begin{figure}
    \centering
    \begin{subfigure}[b]{.92\columnwidth}
        \includegraphics[width=\columnwidth]{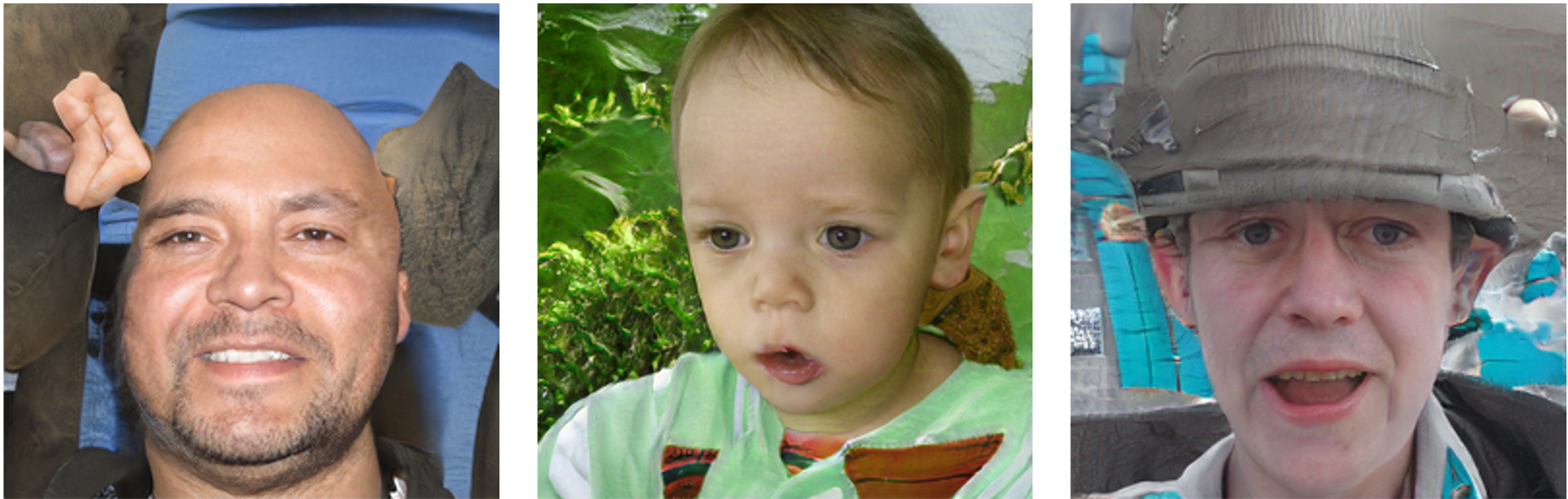}
        \caption{Examples having high AS-i}
        \label{fig:enter-label}
    \end{subfigure}
    \begin{subfigure}[b]{.92\columnwidth}
        \includegraphics[width=\columnwidth]{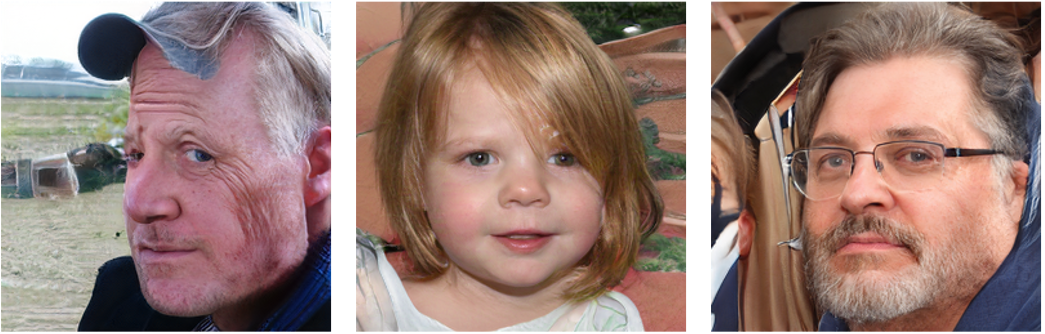}
        \caption{Examples having medium AS-i (around 600)}
        \label{fig:enter-label}
    \end{subfigure}
    \begin{subfigure}[b]{.92\columnwidth}
        \includegraphics[width=\columnwidth]{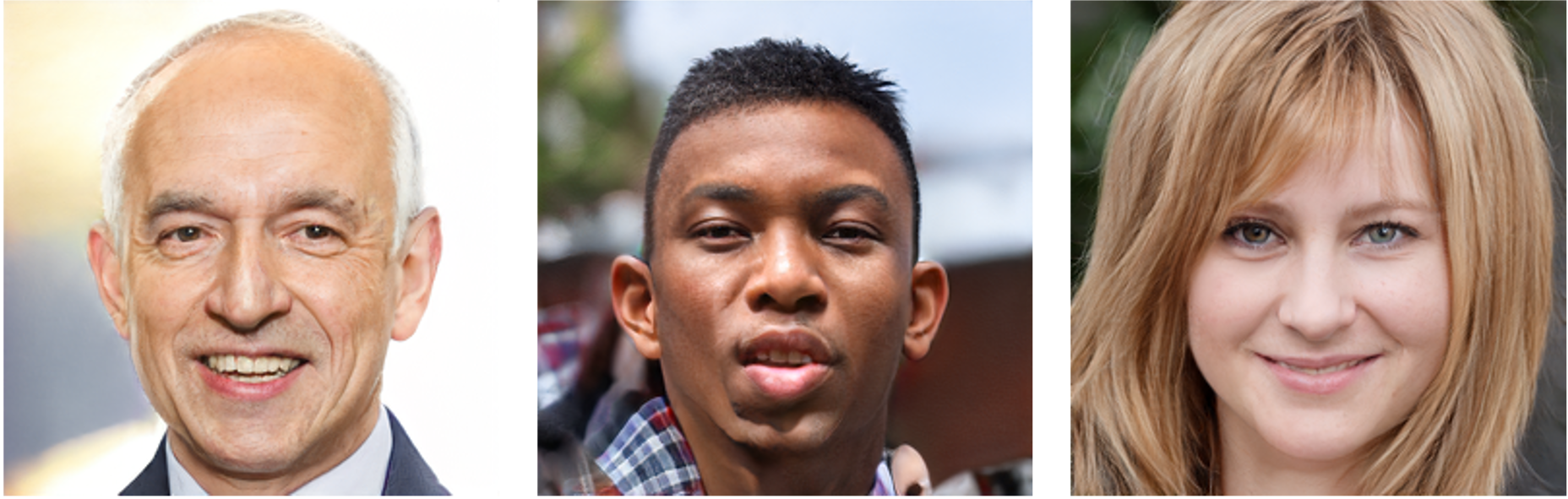}
        \caption{Examples having low AS-i}
        \label{fig:enter-label}
    \end{subfigure}
    \vspace{-1em}
    \caption{\textbf{Examples having various levels of AS-i.}}
    \label{fig:ex-level}
    \vspace{-.5em}
\end{figure}

\noindent \textbf{Experimental settings.}
To verify that AS-i captures human judgement well in terms of the naturalness of images, we conduct a subjective test using the images generated by InsGen~\cite{insgen}.
We set five levels of AS-i (highest in the dataset, 900, 600, 300, and lowest in the dataset) using the ConvNeXt-tiny feature model.
The highest level of AS-i ranges from 1434 to 1746 with an average of 1561 and the lowest level ranges from 53 to 80 with an average of 71.
Then, for each level, we form an image subset by sampling 20 images having AS-i close to the level.
\cref{fig:ex-level} shows several examples of images that belong to the highest, medium (600), and the lowest levels.
14 participants are asked to judge if each image appears natural.
We consider an image to be natural if over 50\% of the participants respond that the image is natural as follows the rule used in \cite{mvt}.
Then, we obtain the proportion of the number of images that are identified as natural in each subset.

\renewcommand{\arraystretch}{1}
\begin{table}[t]
\centering
\small
\begin{tabular}{c|c|cc}
\toprule
AS-i&Human&Rarity \cite{rarity}&Realism \cite{IPR}\\
\midrule
Low&0.75&21.82&1.039\\
300&0.50&18.47&1.036\\
600&0.40&21.51&1.030\\
900&0.30&21.54&1.032\\
high&0.25&20.16&1.010\\
\bottomrule
\end{tabular}
\vspace{-.5em}
\caption{\textbf{Evaluation of AS-i.} For each level of AS-i, the proportion of images judged as natural by participants, the average rarity score, and the average realism score are shown.}
\label{tab: subjective-test}
\vspace{-1.5em}
\centering
\end{table}

\noindent \textbf{Results and comparison.}
\cref{tab: subjective-test} presents the results of the subjective test.
It is observed that as AS-i level increases, fewer images in a subset are judged as natural in a consistent manner.
On the other hand, the average rarity score \cite{rarity} and realism score \cite{IPR} are not aligned well with human perception. 
The PCCs for AS-i, rarity score, and realism score are -0.88, -0.49, and 0.63, respectively.

%% file: sec/6_conclusion.tex
\section{Conclusion}

We have proposed new metrics, AS and AS-i, for evaluation of generative models and individual generated images, respectively. Both are based on \com and \vul, which examine the representation space around the images. \Com captures the curvedness of the representation space, while \vul tests the changes in the representation space under adversarial attack. We demonstrated that the proposed metrics accord well with human judgments and outperform existing metrics.

\vspace{1em}
